\documentclass[sigconf, screen]{acmart}

\AtBeginDocument{%
  }

\setcopyright{acmlicensed}
\copyrightyear{2018}
\acmYear{2018}
\acmDOI{XXXXXXX.XXXXXXX}

\acmConference[Conference acronym 'XX]{Make sure to enter the correct
  conference title from your rights confirmation email}{June 03--05,
  2018}{Woodstock, NY}

\acmISBN{978-1-4503-XXXX-X/2018/06}
\usepackage[ruled,vlined,linesnumbered]{algorithm2e}
\SetKwInput{KwIn}{Require}
\SetKwInput{KwOut}{Output}
\DontPrintSemicolon
\SetAlCapSkip{0.5em}

\usepackage{subcaption}
\usepackage{multirow}
\usepackage{cuted}

\newcommand{\best}[1]{\textbf{#1}}
\usepackage{makecell}
\usepackage{tikz}
\usepackage{colortbl}
\usepackage{placeins}
\SetAlgoCaptionSeparator{.}

\usepackage{booktabs}
\usepackage{array}
\usepackage{graphicx}
\usepackage{adjustbox}
\usepackage{threeparttable}
\usepackage{dblfloatfix}
\usepackage{enumitem}
\definecolor{modelbg}{RGB}{235,242,250}  
\DeclareCaptionFont{mysubcap}{\fontsize{7}{11}\selectfont\fontfamily{qhv}\bfseries}
\newcommand{\method}{DTR}

\renewcommand\footnotetextcopyrightpermission[1]{}
\renewcommand{\arraystretch}{1.04}
\begin{document}

\title{Relaxing Anchor-Frame Dominance for Mitigating Hallucinations in Video Large Language Models}

\author{Zijian Liu}
\affiliation{%
  \institution{University of Electronic Science and Technology of China}
  \city{Chengdu}
  \country{China}
}
\email{2023080907011@std.uestc.edu.cn}

\author{Sihan Cao}
\affiliation{%
  \institution{University of Electronic Science and Technology of China}
  \city{Chengdu}
  \country{China}
}
\email{2023080903002@std.uestc.edu.cn}

\author{Pengcheng Zheng}
\affiliation{%
  \institution{University of Electronic Science and Technology of China}
  \city{Chengdu}
  \country{China}
}
\email{zpc777@std.uestc.edu.cn}

\author{Kuien Liu}
\affiliation{%
  \institution{Institute of Software Chinese Academy of Sciences}
  \city{Beijing}
  \country{China}
}
\email{kuien@iscas.ac.cn}

\author{Caiyan Qin}
\affiliation{%
  \institution{Harbin Institute of Technology, Shenzhen}
  \city{Shenzhen}
  \country{China}
}
\email{qincaiyan@hit.edu.cn}

\author{Xiaolin Qin}
\affiliation{%
  \institution{Chengdu Institute of Computer Applications, Chinese Academy of Sciences, University of the Chinese Academy of Sciences}
  \city{Chengdu}
  \country{China}
}
\email{qinxl2001@126.com}

\author{Jiwei Wei}
\affiliation{%
  \institution{University of Electronic Science and Technology of China}
  \city{Chengdu}
  \country{China}
}
\email{weijiwei@uestc.edu.cn}

\author{Chaoning Zhang}
\affiliation{%
  \institution{University of Electronic Science and Technology of China}
  \city{Chengdu}
  \country{China}
}
\email{chaoningzhang1990@gmail.com}

\begin{abstract}
Recent Video Large Language Models (Video-LLMs) have demonstrated strong capability in video understanding, yet they still suffer from hallucinations. Existing mitigation methods typically rely on training, input modification, auxiliary guidance, or additional decoding procedures, while largely overlooking a more fundamental challenge. During generation, Video-LLMs tend to over-rely on a limited portion of temporal evidence, leading to temporally imbalanced evidence aggregation across the video. To address this issue, we investigate a decoder-side phenomenon in which the model exhibits a temporally imbalanced concentration pattern. We term the frame with the highest aggregated frame-level attention mass the anchor frame. We find that this bias is largely independent of the input video and instead appears to reflect a persistent, model-specific structural or positional bias, whose over-dominance is closely associated with hallucination-prone generation. Motivated by this insight, we propose Decoder-side Temporal Rebalancing (DTR), a training-free, layer-selective inference method that rebalances temporal evidence allocation in middle-to-late decoder layers without altering visual encoding or requiring auxiliary models. DTR adaptively calibrates decoder-side visual attention to alleviate temporally imbalanced concentration and encourage under-attended frames to contribute more effectively to response generation. In this way, DTR guides the decoder to ground its outputs in temporally broader and more balanced video evidence. Extensive experiments on hallucination and video understanding benchmarks show that DTR consistently improves hallucination robustness across diverse Video-LLM families, while preserving competitive video understanding performance and high inference efficiency. The code is publicly available to support reproducibility at \url{https://github.com/ifwub234/DTR}.

\end{abstract}


\maketitle

\begin{figure*}[t]
    \captionsetup[subfigure]{labelfont=mysubcap}
    \centering

    \begin{subfigure}[t]{0.498\textwidth}
        \centering
        \includegraphics[width=\linewidth]{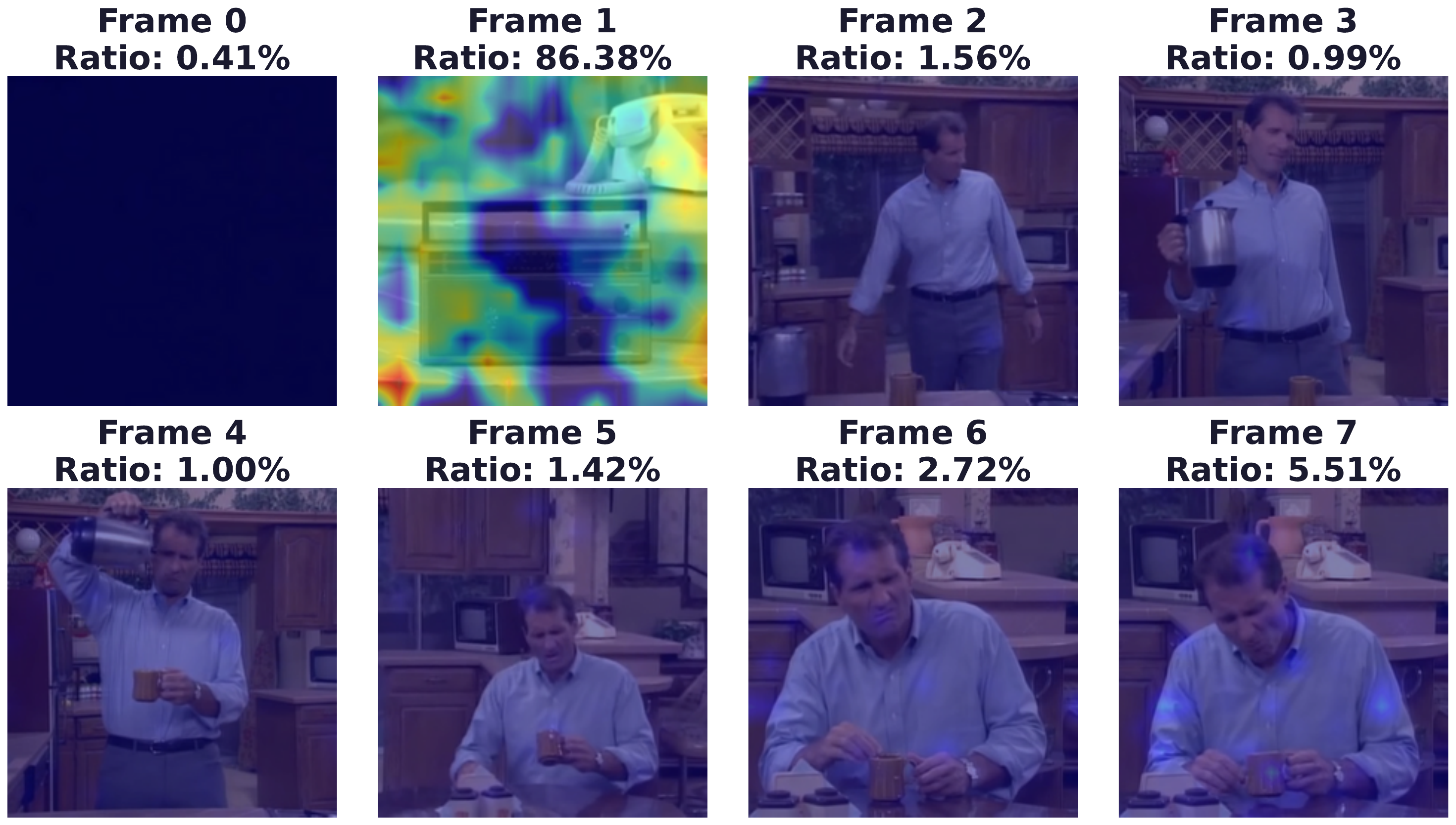}

        \vspace{0.0pt}
        {\centering
        \begin{minipage}{0.96\linewidth}
        \centering
        \fontsize{7}{11}\selectfont\fontfamily{qhv}\bfseries
        Q: Does 'he walks into the kitchen and pours a cup of coffee as he gloats' happen earlier than 'an old style radio is sitting on a counter'?\\
        A: Yes \qquad GT: No
        \end{minipage}
        \par}
        \caption{\fontsize{7}{11}\selectfont\fontfamily{qhv}\bfseries
        Video-LLaVA with an early anchor frame.}
        \label{fig:intro_videollava}
    \end{subfigure}
    \begin{subfigure}[t]{0.498\textwidth}
        \centering
        \includegraphics[width=\linewidth]{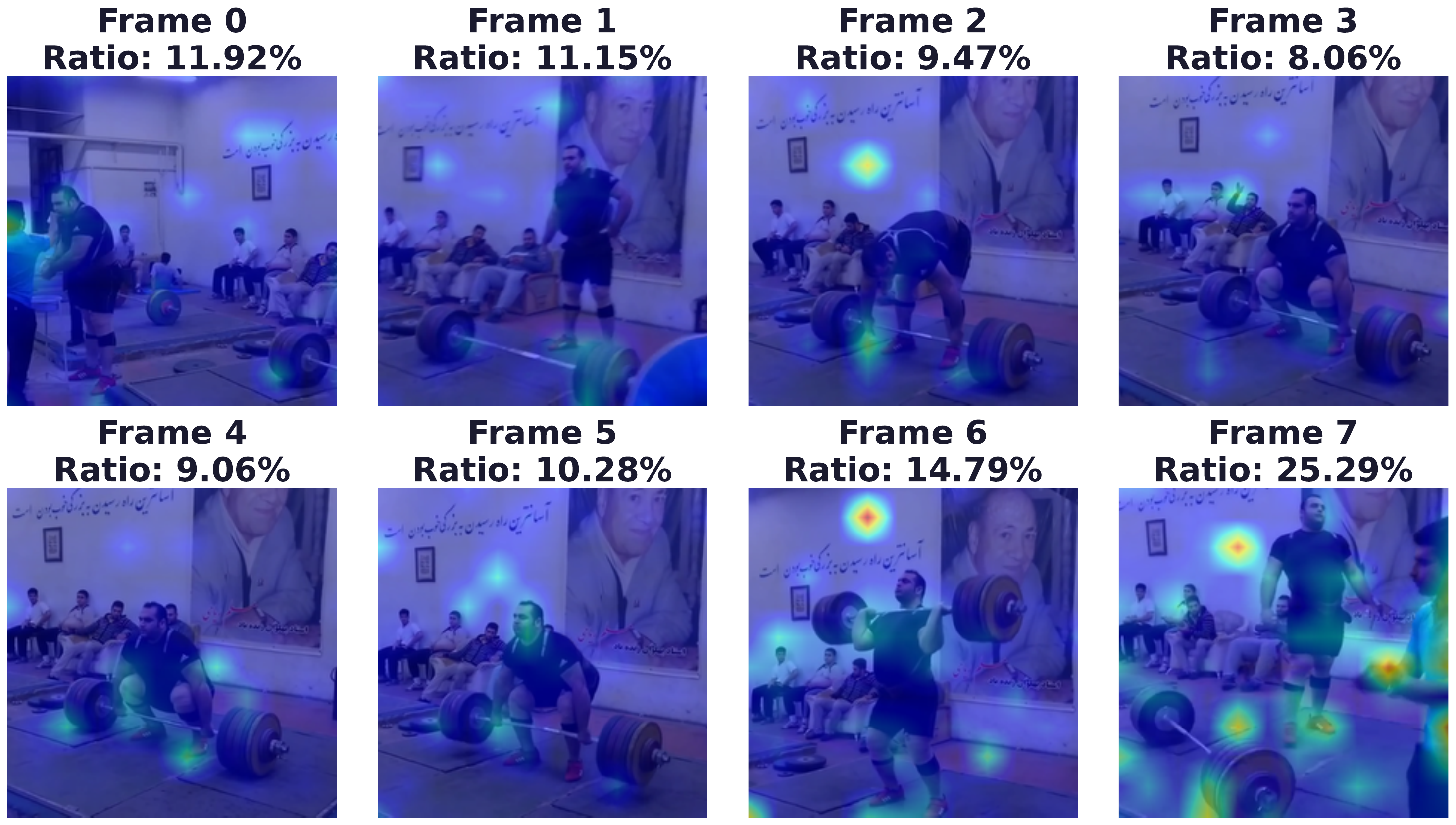}

        \vspace{0.0pt}
        {\centering
        \begin{minipage}{0.96\linewidth}
        \centering
        \fontsize{7}{11}\selectfont\fontfamily{qhv}\bfseries
        Q: Does 'a man puts chalk on his hands' happen later than 'he bends down and lifts a large weight over his head'?\\
        A: Yes \qquad GT: No
        \end{minipage}
        \par}
        \caption{\fontsize{7}{11}\selectfont\fontfamily{qhv}\bfseries
        LLaVA-NeXT-Video with a late anchor frame.}
        \label{fig:intro_llavanext}
    \end{subfigure}

    \caption{Anchor-frame dominance in two Video-LLMs. Video-LLaVA focuses on frame 1, while LLaVA-NeXT-Video focuses on frame 7, leading to incorrect temporal judgments.}
    \vspace{1.85mm}
    \label{fig:intro_anchor_examples}
\end{figure*}

\section{Introduction}
Recent Video Large Language Models (Video-LLMs), such as Video-LLaVA \cite{lin2024video}, LLaVA-NeXT-Video \cite{zhang2024llavanextvideo}, Qwen2.5-VL \cite{bai2025qwen25vltechnicalreport}, Qwen3-VL \cite{bai2025qwen3} and InternVL3.5 \cite{wang2025internvl3}, have achieved remarkable progress in video question answering, captioning, and open-ended video understanding \cite{tang2025video,fu2025video,maaz2024video,li2025videochat,zhang2023video,cheng2024videollama}. Despite these advances, they still suffer from hallucination, where the generated response is not supported by the visual input or even contradicts the visual content \cite{huang2024visual}. This issue substantially undermines the reliability of Video-LLMs in real-world applications, especially in safety-critical scenarios such as autonomous driving \cite{ding2024holistic} and healthcare \cite{lin2025has}. Therefore, mitigating hallucination is a key step toward faithful and trustworthy multimodal systems.

Hallucination mitigation has been extensively studied in image-based LLMs, with representative approaches spanning training-based alignment, post-hoc correction, training-free decoding-time intervention, and internal representation modification, as exemplified by M-HalDetect/FDPO \cite{gunjal2024detecting}, RLHF-V \cite{yu2024rlhf}, LURE \cite{zhou2023analyzing}, Woodpecker \cite{yin2024woodpecker}, VCD \cite{leng2024mitigating}, ICD \cite{wang2024mitigating}, M3ID \cite{favero2024multi}, Interpreting and Editing Vision-Language Representations \cite{jiang2024interpreting}, Look Twice \cite{zou2024look}, and ClearSight \cite{yin2025clearsight}. However, directly extending these ideas to videos is nontrivial. Unlike image hallucination, which is primarily a static visual grounding problem, video hallucination additionally requires models to capture temporal dependencies, motion transitions, and event evolution across frames. Benchmarks such as EventHallusion \cite{zhang2024eventhallusion}, VideoHallucer \cite{wang2024videohallucer}, VidHal \cite{choong2024vidhal}, and VideoHallu \cite{li2025videohallu} show that video hallucination is not only about grounding visual content, but also about reasoning over temporally distributed evidence. As a result, many image-oriented mitigation strategies do not explicitly address how temporal evidence should be allocated and aggregated during generation. Existing mitigation methods for Video-LLMs can be broadly grouped into two paradigms: \emph{training-based} methods, which improve faithfulness by adapting the model through tuning or preference optimization, such as VISTA-LLAMA \cite{ma2024vista}, MASH-VLM \cite{bae2025mash}, HAVEN \cite{gao2025exploring}, and PaMi-VDPO \cite{ding2025pami}, and \emph{training-free} methods, which instead intervene at inference time through contrastive decoding or feature calibration, such as TCD \cite{zhang2024eventhallusion}, SEASON \cite{wu2025season}, MotionCD \cite{kong2025mhbench}, and DINO-HEAL \cite{li2025vidhalluc}.

However, despite the steady progress made in recent training-based and training-free studies on video hallucination mitigation, existing methods still face notable limitations. Training-based methods can improve hallucination mitigation, but they usually incur additional optimization cost and are not explicitly designed to regulate how temporally distributed evidence is allocated during generation. Training-free methods are more flexible for deployment, yet many still rely on input modification, auxiliary guidance, or extra decoding procedures, which increase inference complexity and computational overhead. More importantly, both paradigms still struggle to encourage Video-LLMs to fully exploit temporal evidence throughout generation, as the decoder often exhibits a stable yet temporally imbalanced concentration pattern, causing evidence to be concentrated on limited temporal content while broader video information remains under-utilized.

To address this issue, we revisit video hallucination through the lens of decoder-side temporal evidence imbalance. As illustrated in Figure~\ref{fig:intro_anchor_examples}, aggregating decoder attention over video frames reveals that generation often exhibits a temporally imbalanced concentration pattern, whose highest-attention frame under the aggregated frame-level attention statistic is termed the anchor frame. Importantly, this concentration pattern is largely independent of the input video and instead appears to reflect a model-specific, content-agnostic bias. Figure~\ref{fig:intro_anchor_examples} further shows how such over-concentration can distort temporal reasoning: Video-LLaVA concentrates most of its attention on the early frame containing the old-style radio while overlooking the subsequent coffee-pouring action, whereas LLaVA-NeXT-Video over-focuses on late lifting frames while underweighting earlier frames showing the chalking action and overlooking the event in which he bends down and lifts a large weight over his head, leading both models to incorrect temporal judgments. More importantly, our subsequent analyses and experiments further confirm that this concentration pattern is closely associated with hallucination, as over-reliance on limited temporal evidence suppresses complementary cues from the broader video context and results in temporally biased evidence aggregation during generation.

Motivated by this insight, together with our observation that broader video evidence becomes increasingly under-utilized in middle-to-late decoder layers, we propose Decoder-side Temporal Rebalancing (DTR), a simple, training-free, and layer-selective decoder-side intervention for mitigating hallucination by rebalancing temporal evidence during generation. Concretely, DTR operates on the decoder’s temporal attention pattern in selected middle-to-late layers, alleviating temporally imbalanced concentration while compensating under-attended frames in a broad and adaptive manner. 
As a lightweight plug-in, DTR requires no retraining, no modification to the visual encoder or input video, 
and no extra decoding branches, making it easy to integrate into existing Video-LLMs.

Our main contributions are summarized as follows:

\begin{itemize}[leftmargin=1.2em,itemsep=2pt,topsep=2pt,parsep=0pt,partopsep=0pt]

    \item We identify and empirically validate anchor-frame dominance, a persistent decoder-side temporal bias in Video-LLMs that is largely content-agnostic, remains stable under diverse inputs and frame-sampling settings, suppresses the utilization of broader temporal evidence, and is closely associated with hallucination-prone generation behavior in practice.

    \item We propose \emph{Decoder-side Temporal Rebalancing (DTR)}, a simple, lightweight, and training-free inference method that mitigates hallucinations by rebalancing temporally distributed visual evidence against anchor-frame dominance during decoding.

    \item Extensive experiments on hallucination mitigation and general video understanding benchmarks demonstrate that DTR consistently improves response faithfulness across diverse Video-LLM families, while preserving competitive general video understanding performance and favorable decoding efficiency.
\end{itemize}

\section{Related works}

\subsection{Hallucination Mitigation in Image-LLMs}
Hallucination mitigation in image-based LLMs has been studied much more extensively, and existing methods can be broadly grouped into four categories. First, training-based approaches improve visual grounding through additional supervision, preference alignment, or continual fine-tuning, as exemplified by nSFT \cite{zhu2025continual}, RLHF-V \cite{yu2024rlhf}, and M-HalDetect/FDPO \cite{gunjal2024detecting}. Second, post-hoc correction methods revise an initial response after generation through visual verification or rewriting pipelines, as in Woodpecker \cite{yin2024woodpecker} and LURE \cite{zhou2023analyzing}. Third, training-free decoding-time methods mitigate hallucinations by directly modifying the decoding distribution, such as VCD \cite{leng2024mitigating} and ICD \cite{wang2024mitigating} through contrastive decoding, and M3ID \cite{favero2024multi} by increasing visual prompt dependency during generation. Fourth, attention- or representation-level intervention methods improve faithfulness by enhancing visual evidence or directly editing intermediate representations, as explored by PAI \cite{liu2024paying} and ClearSight \cite{yin2025clearsight} through attention redistribution, and by Interpreting and Editing Vision-Language Representations \cite{jiang2024interpreting} and Look Twice \cite{zou2024look} through latent editing or visual re-injection. However, most image-oriented methods still target static spatial hallucinations, where the core problem is object or attribute grounding, rather than the allocation of temporal evidence across frames in videos.

\subsection{Hallucination Mitigation in Video-LLMs}

Compared with image hallucination, video hallucination is not merely a temporal extension of visual grounding, because Video-LLMs must additionally model temporal dynamics, event order, and temporal faithfulness \cite{gao2025exploring,wu2025season}. As a result, image-side mitigation strategies do not directly resolve how evidence should be aggregated across frames during generation \cite{wu2025season}. Existing methods for Video-LLMs can be broadly divided into training-based and training-free approaches. Training-based methods typically mitigate hallucinations through fine-tuning, preference optimization, or architectural redesign. Representative examples include HAVEN \cite{gao2025exploring}, MASH-VLM \cite{bae2025mash}, VISTA-LLAMA \cite{ma2024vista}, and PaMi-VDPO \cite{ding2025pami}. There is also work that introduces activation engineering for Video-LLM hallucination mitigation \cite{cai2025mitigating}. In contrast, training-free methods intervene at test time, including Temporal Contrastive Decoding (TCD) \cite{zhang2024eventhallusion}, which suppresses event priors by contrasting outputs from original and temporally degraded videos. Self-Diagnostic Contrastive Decoding (SEASON) \cite{wu2025season}, which uses temporally homogenized negatives and token-wise self-diagnostic weights to penalize temporal and spatial hallucinations. MotionCD \cite{kong2025mhbench} mitigates motion hallucination by contrasting outputs from the original and reversed videos. DINO-HEAL \cite{li2025vidhalluc}, which leverages DINOv2 \cite{oquab2023dinov2} saliency to reweight visual features toward critical regions. In contrast, our method addresses a decoder-side temporal imbalance by selectively rebalancing temporally distributed visual evidence in middle-to-late decoder layers, so that generation is less driven by biased concentration on limited temporal evidence and better grounded in the broader video context.

\begin{figure}[t]
    \centering
    \includegraphics[width=\columnwidth]{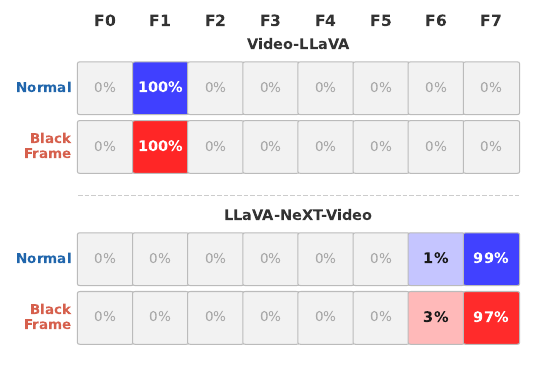}

    \caption{Anchor-frame behavior under black-frame intervention. Blue indicates the anchor-frame distribution under normal video inference, and red indicates the distribution after replacing the anchor frame with an all-black frame before visual encoding. In both Video-LLaVA and LLaVA-NeXT-Video, the dominant temporal position remains unchanged. Percentages denote how often each frame is selected as the anchor frame on the Object-Relation and Temporal subsets of VideoHallucer.}

    \label{fig:visualize_hall}
\end{figure}
\section{Method}
\label{sec:method}

\subsection{Preliminary Observation: Anchor-Frame}
\label{sec:preliminary_observation}

To quantify a simple but consistent empirical observation across multiple Video-LLMs, we study a phenomenon in which decoder attention exhibits a temporally imbalanced concentration pattern during generation. We use the anchor frame to denote the frame with the highest aggregated frame-level attention mass, which serves as a compact summary of the dominant temporal focus under this pattern. This concentration pattern is largely independent of the video content and instead reflects a model-specific structural or positional bias.

To keep the notation consistent with the method section, we use \(\tau\) to denote the current forward state: \(\tau=\mathrm{pre}\) corresponds to the prefill pass, while \(\tau=t\) corresponds to the \(t\)-th autoregressive decoding step. In this preliminary analysis, we consider only the prefill state \(\tau=\mathrm{pre}\). Let \(\mathcal{Q}_{\mathrm{pre}}\) denote the set of textual query positions after the visual token block, and let \(z_{q,j}^{(l,h)}\) denote the raw attention logit at decoder layer \(l\) and head \(h\), from query position \(q \in \mathcal{Q}_{\mathrm{pre}}\) to key token \(j\), where \(H\) is the number of attention heads. We first average the raw logits across all collected query positions and attention heads:
\begin{equation}
\bar{z}_{\mathrm{pre},j}^{(l)}
=
\frac{1}{H |\mathcal{Q}_{\mathrm{pre}}|}
\sum_{q \in \mathcal{Q}_{\mathrm{pre}}}
\sum_{h=1}^{H}
z_{q,j}^{(l,h)}.
\label{eq:avg_logit_pre}
\end{equation}
Let \(\mathcal{V}_i\) denote the set of visual tokens belonging to frame \(i\). We then define the frame-level attention mass of frame \(i\) at layer \(l\) as
\begin{equation}
a_{\mathrm{pre},i}^{(l)}
=
\sum_{j \in \mathcal{V}_i}
\operatorname{Softmax}_j
\!\left(
\bar{z}_{\mathrm{pre},j}^{(l)}
\right),
\label{eq:frame_score}
\end{equation}
where \(\operatorname{Softmax}_j(\cdot)\) is taken over the full key dimension \(j\). The anchor frame of the sample is then defined by averaging frame-level attention masses across the analyzed layer set \(\mathcal{L}\):

\begin{equation}
\kappa
=
\arg\max_i
\frac{1}{|\mathcal{L}|}
\sum_{l \in \mathcal{L}} a_{\mathrm{pre},i}^{(l)}.
\label{eq:anchor_frame}
\end{equation}
Here, \(a_{\mathrm{pre},i}^{(l)}\) and \(\kappa\) are introduced only for offline statistical analysis of anchor-frame bias. In contrast, the inference-time DTR method below uses a stage-dependent query set \(\mathcal{Q}_\tau\) for score estimation and a single query position \(q_\tau^\star\) for actual logit modification.

\begin{figure}[!t]
    \centering
    \includegraphics[width=\linewidth]{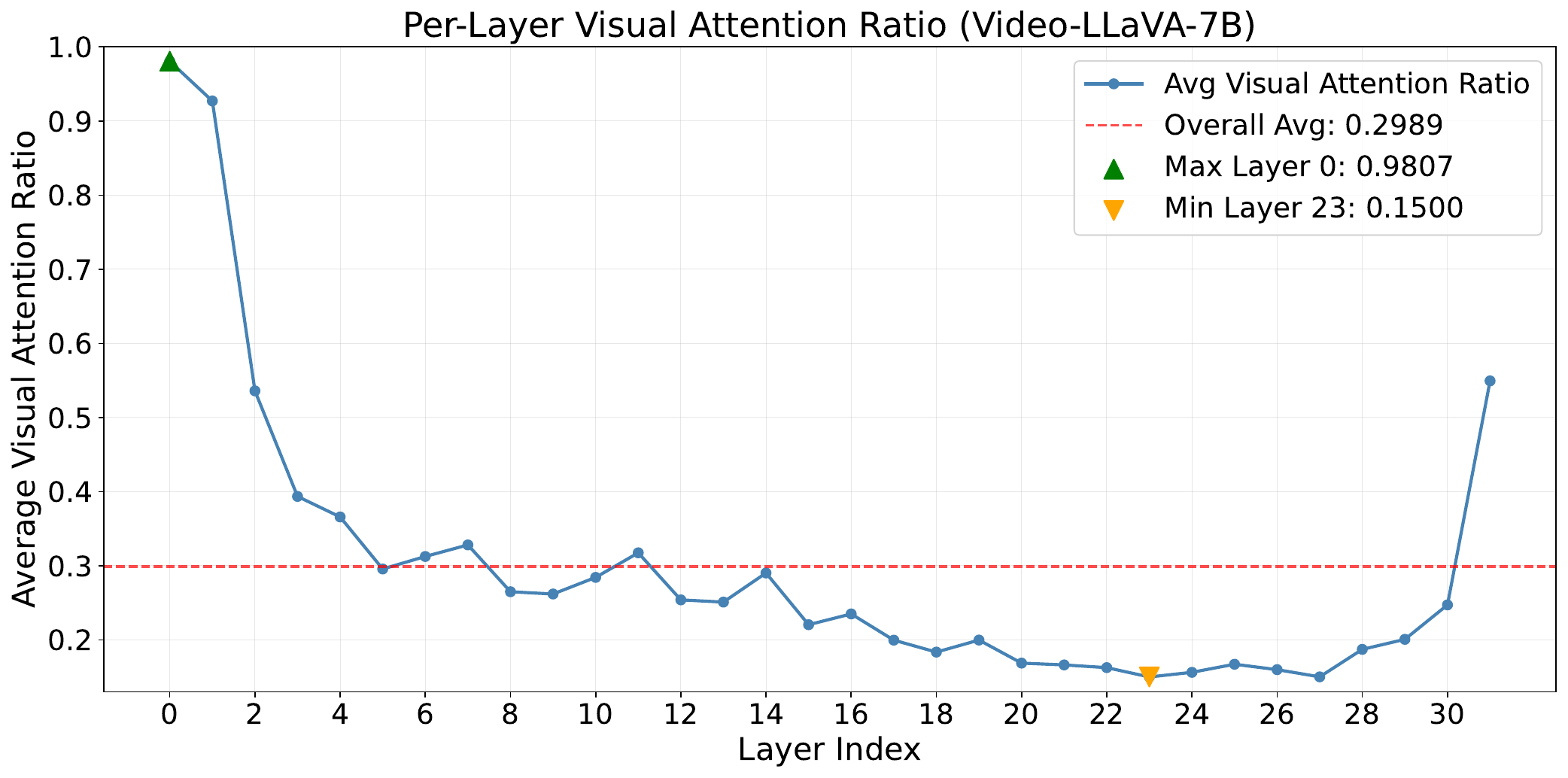}
    \caption{Layer-wise visual attention ratio in Video-LLaVA-7B. The curve shows the average visual attention ratio across decoder layers over samples from the Object-Relation and Temporal subsets of
VideoHallucer. Visual attention remains low across most decoder layers, particularly in the middle-to-late range.}
    \label{fig:visual_ratio}
\end{figure}

Our analysis yields three recurring findings. First, anchor-frame dominance is highly stable and largely content-agnostic. Across different samples from the same model, the location of $\kappa$ remains remarkably consistent. This pattern also persists under a black-frame intervention: when we replace the anchor frame with an all-black frame before visual encoding, the model still concentrates most of its attention on the same temporal position, as shown in Figure~\ref{fig:visualize_hall}. Similar anchor-selection bias beyond the LLaVA family is further analyzed in Appendix~\ref{app:add_anchor_stats}.
 Together, these results suggest that anchor-frame dominance is not simply driven by the visual semantics of a particular frame, but instead reflects a model-specific structural or positional bias, rather than an adaptive focus on the most informative visual evidence in each video. We further verify that this bias persists under denser frame sampling. Additional qualitative visualizations and black-frame analysis with varying frame sampling are provided in Appendix~\ref{app:frame_sampling}.
 Second, the decoder allocates limited attention to visual tokens across layers, even though visual tokens dominate the input length. As illustrated in Figure~\ref{fig:visual_ratio}, the layer-wise visual attention ratio remains low overall and stays below the overall average throughout a sustained middle-to-late range. Under anchor-frame dominance, this weak visual reliance further suppresses visual evidence from non-dominant frames. As a result, broader temporal evidence is increasingly neglected in later decoding stages, making the model more vulnerable to language-prior-driven hallucination. Third, the anchor frame appears to be closely associated with hallucination-prone decoding, rather than simply playing a positive role in visual grounding. To further probe this connection, we conduct a masking experiment on LLaVA-NeXT-Video-7B over VideoHallucer. All video frames are first encoded normally by the visual encoder, and masking is applied only in decoder layers 1--31 by setting the attention logits of the selected frame tokens to $-\infty$. As shown in Table~\ref{tab:anchor_mask}, masking only the anchor frame improves the overall accuracy from $27.12\%$ to $29.29\%$, whereas randomly masking one frame reduces it to $26.61\%$, and masking a non-anchor frame further degrades it to $25.30\%$. This comparison suggests that the anchor frame mainly functions as an over-concentrated temporal node that attracts excessive decoder attention and suppresses contributions from other frames. These findings further support that hallucination in Video-LLMs is closely related to temporally imbalanced evidence aggregation during decoding. This motivates us to rebalance temporal evidence usage at inference time, especially in the middle-to-late decoder layers, without modifying the visual encoder or retraining the model.

\begin{table}[!htbp]
\caption{Masking study on VideoHallucer (\textit{All} accuracy, \%) using LLaVA-NeXT-Video-7B. All video frames are first encoded normally by the visual encoder, and masking is applied only to the selected frame tokens in decoder layers 1--31.}
\label{tab:anchor_mask}
\centering
\begin{tabular*}{\columnwidth}{@{\extracolsep{\fill}}ccc@{}}
\toprule
Setting & All Acc. (\%) & $\Delta$ \\
\midrule
Normal inference & 27.12 & 0.00 \\
Mask anchor frame only & \textbf{29.29} & \textbf{+2.17} \\
Randomly mask one frame & 26.61 & -0.51 \\
Mask Frame 4 (non-anchor) & 25.30 & -1.82 \\
\bottomrule
\end{tabular*}
\end{table}
\begin{figure*}[!t]
\centering
\includegraphics[width=\textwidth]{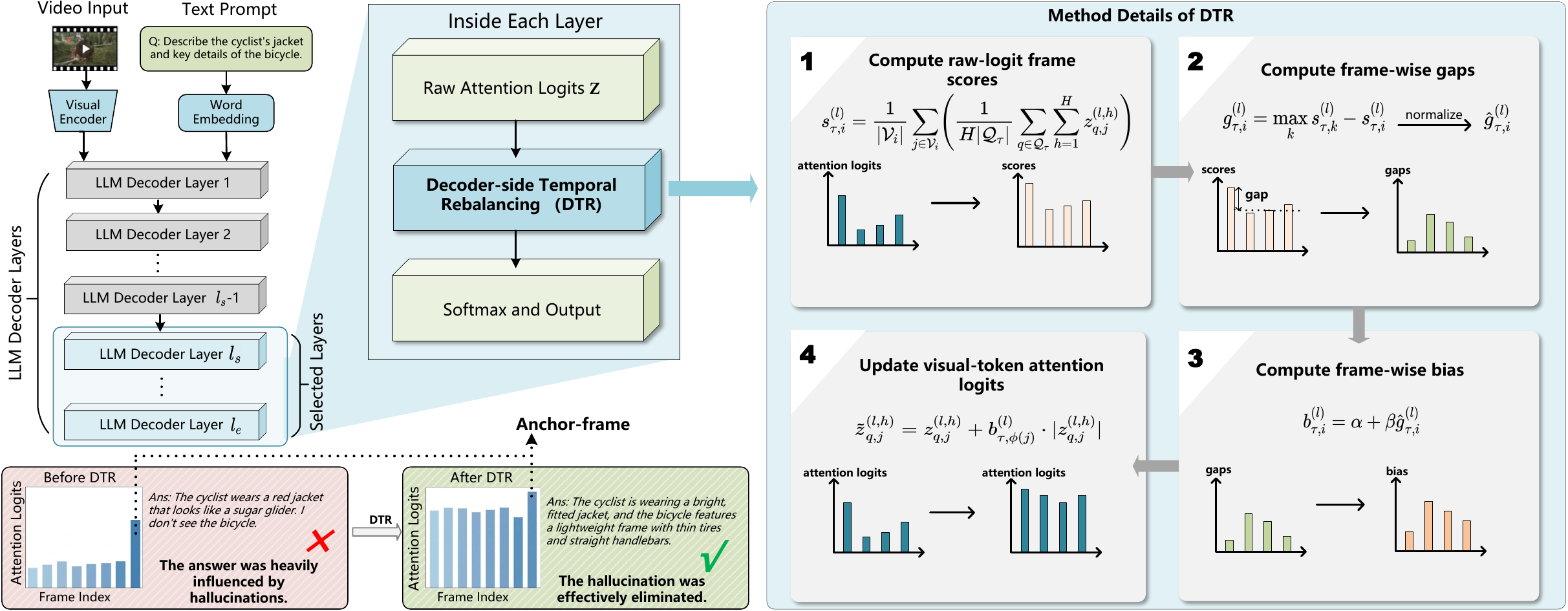}
\caption{Overview of the proposed decoder-side temporal rebalancing (DTR) framework. Given a video and a text prompt, DTR operates in selected middle-to-late decoder layers by computing frame-level scores from raw attention logits, estimating each frame’s gap to the current dominant frame, converting the normalized gaps into positive frame-wise biases, and injecting the biases into visual-token attention logits before softmax. The intervention is global and relative: it preserves all frames while assigning larger compensation to under-attended ones, thereby reducing anchor-frame over-dominance and improving temporal grounding.}
\label{fig:DTR}
\end{figure*}
\subsection{Decoder-side Temporal Rebalancing}
\label{sec:dtr}

Building on these observations, we propose Decoder-side Temporal Rebalancing (DTR), an inference-time method for rebalancing temporal evidence during decoding. Our design is motivated by two coupled phenomena: the decoder exhibits anchor-frame dominance, while weak visual reliance in middle-to-late layers further suppresses evidence outside the dominant temporal focus. Accordingly, DTR intervenes directly on decoder attention, enhances under-attended frames in a relative and global manner, and applies the modification only to middle-to-late decoder layers. Figure~\ref{fig:DTR} illustrates the framework.

Given an input video \(V\), we uniformly sample \(N\) frames and feed them into the visual encoder. After visual tokenization and projection, the video is represented as
\begin{equation}
\mathcal{X}^{v} = \{\mathcal{V}_1, \mathcal{V}_2, \dots, \mathcal{V}_N\},
\end{equation}
where \(\mathcal{V}_i = \{v_{i,1}, v_{i,2}, \dots, v_{i,M_i}\}\) denotes the visual tokens of frame \(i\). These visual tokens are concatenated with textual tokens and processed causally by the LLM decoder.

For notation, let \(\tau\) denote the current forward state: \(\tau=\mathrm{pre}\) for the prefill pass, and \(\tau=t\) for the \(t\)-th autoregressive decoding step. Let \(\mathcal{Q}_\tau\) denote the set of query positions used only for score estimation, and let \(q_\tau^\star\) denote the single query position whose logits are actually modified. Concretely, during prefill, \(\mathcal{Q}_{\mathrm{pre}}\) is the set of textual query positions after the visual token block, and \(q_{\mathrm{pre}}^\star\) is always the last query position of the current prefill pass. During autoregressive decoding at step \(t\), we use \(\mathcal{Q}_t=\{t\}\) and \(q_t^\star=t\). Let \(z_{q,j}^{(l,h)}\) denote the raw self-attention logit at decoder layer \(l\) and head \(h\), from query position \(q\) to key position \(j\), after applying the standard causal mask but before softmax.

First, we compute a frame-level score directly from the raw attention logits:
\begin{equation}
s_{\tau,i}^{(l)}
=
\frac{1}{|\mathcal{V}_i|}
\sum_{j \in \mathcal{V}_i}
\left(
\frac{1}{H|\mathcal{Q}_\tau|}
\sum_{q \in \mathcal{Q}_\tau}
\sum_{h=1}^{H}
z_{q,j}^{(l,h)}
\right),
\label{eq:raw_frame_score}
\end{equation}
where \(H\) is the number of attention heads. Then, we estimate how far each frame is from the currently dominant one by
\begin{equation}
g_{\tau,i}^{(l)}
=
\max_k s_{\tau,k}^{(l)} - s_{\tau,i}^{(l)},
\label{eq:score_gap}
\end{equation}
and normalize the gap as
\begin{equation}
\hat{g}_{\tau,i}^{(l)}
=
\frac{g_{\tau,i}^{(l)}}{\max_k g_{\tau,k}^{(l)} + \epsilon},
\label{eq:norm_gap}
\end{equation}
where \(\epsilon = 10^{-6}\) is a small constant for numerical stability. Next, we convert the normalized deficit into a positive frame-wise bias,
\begin{equation}
b_{\tau,i}^{(l)} = \alpha + \beta \hat{g}_{\tau,i}^{(l)},
\label{eq:frame_bias}
\end{equation}
where \(\alpha \ge 0\) is a shared adjustment term applied uniformly to all visual frames, and \(\beta \ge 0\) controls the additional compensation assigned to under-scored frames. In this way, the method preserves the participation of all frames in temporal evidence reallocation, while giving relatively stronger adjustment to under-attended frames to counteract anchor-frame over-dominance and encourage a more balanced multi-frame grounding pattern.

Finally, let \(\phi(j)\) denote the frame index of visual token \(j\). We apply the bias to visual-token attention logits before softmax, and only within a selected layer range \(\mathcal{L}^{\star}=\{l_s, l_s+1, \dots, l_e\}\):
\begin{equation}
\tilde{z}_{q,j}^{(l,h)}
=
\begin{cases}
z_{q,j}^{(l,h)} + b_{\tau,\phi(j)}^{(l)} \cdot |z_{q,j}^{(l,h)}|,
& \text{if } l \in \mathcal{L}^{\star},\ q=q_\tau^\star,\ j \in \mathcal{X}^{v}, \\[4pt]
z_{q,j}^{(l,h)},
& \text{otherwise},
\end{cases}
\label{eq:full_update}
\end{equation}
and the final attention weights are computed as
\begin{equation}
\tilde{A}_{q,j}^{(l,h)}
=
\mathrm{Softmax}\!\left(\tilde{z}_{q,j}^{(l,h)}\right).
\label{eq:modified_attention}
\end{equation}

This design is global and relative: it rebalances contributions from all frames while assigning larger compensation to under-attended ones. We restrict the intervention to middle-to-late decoder layers because Figure~\ref{fig:visual_ratio} shows that visual attention becomes particularly weak in these layers, often remaining below the overall average for a sustained range. Under anchor-frame dominance, this weak visual reliance further suppresses broader grounded evidence, making middle-to-late layers the most critical stage for correcting temporally imbalanced evidence aggregation while avoiding unnecessary disruption to early multimodal processing.

\begin{table*}[t]
\caption{Results on LLaVA-based video models, including LLaVA-NeXT-Video-7B and Video-LLaVA-7B, across EventHallusion, VideoHallucer, and MVBench. Higher is better, and all values are reported as accuracy (\%). Bold indicates the best result.}
\label{tab:llava_based_models}
\centering
\small
\renewcommand{\arraystretch}{1.08}
\resizebox{\textwidth}{!}{
\begin{tabular}{cccccccccccc}
\toprule
\multirow{2}{*}{\raisebox{-0.3ex}{Method}}
& \multicolumn{4}{c}{EventHallusion~\cite{zhang2024eventhallusion}}
& \multicolumn{6}{c}{VideoHallucer~\cite{wang2024videohallucer}}
& MVBench~\cite{li2024mvbench} \\
\cmidrule(lr){2-5}
\cmidrule(lr){6-11}
\cmidrule(lr){12-12}
& Entire & Mix & Misleading & Overall
& Object-Relation & Temporal & Semantic Detail & Factual & Non-factual & Overall
& Overall \\
\midrule
\rowcolor{ACMBlue!10}
\multicolumn{12}{c}{\textbf{LLaVA-NeXT-Video-7B}~\cite{zhang2024llavanextvideo}} \\
\midrule
Baseline \cite{zhang2024llavanextvideo}    & 44.74 & 68.39 & 50.00 & 57.21 & 52.00 & 17.61 & 37.00 & 3.00 & 26.00 & 27.12 & 43.575 \\
+TCD \cite{zhang2024eventhallusion}        & 45.61 & \best{69.43} & 52.94 & 58.68 & 51.00 & 14.20 & 36.00 & 1.50 & 22.50 & 25.04 & 43.77  \\
+VCD \cite{leng2024mitigating}  & 42.11 & 63.73 & 51.96 & 54.77 & 45.50 & \best{29.55} & 39.50 & \best{19.50} & \best{38.50} & \best{34.51} & 43.53 \\
+DINO-HEAL \cite{li2025vidhalluc}  & 47.37 & 67.88 & 50.98 & 57.95 & 51.00 & 18.75 & 37.50 & 3.50 & 27.00 & 27.55 & 43.65 \\
\textbf{+DTR (Ours)}       & \best{51.75} & 64.25 & \best{66.67} & \best{61.37} & \best{53.50} & 28.98 & \best{44.50} & 4.00 & 31.00 & 32.40 & \best{45.07} \\

\midrule
\rowcolor{ACMBlue!10}
\multicolumn{12}{c}{\textbf{Video-LLaVA-7B}~\cite{lin2024video}} \\
\midrule
Baseline \cite{lin2024video}    & 28.95 & 58.03 & 41.18 & 45.72 & 34.00 & 8.52 & 11.50 & 3.50 & 27.00 & 16.90 & 43.90 \\
+TCD \cite{zhang2024eventhallusion}       & 31.58 & \best{68.91} & 40.20 & 51.34 & 36.00 & 19.32 & 12.50 & 5.00 & 28.00 & 20.16 & \best{44.73} \\
+VCD \cite{leng2024mitigating}  & 40.35 & 52.85 & 39.22 & 45.97 & 31.00 & 26.70 & 25.00 & \best{20.00} & 29.00 & 26.34 & 41.77 \\
+DINO-HEAL \cite{li2025vidhalluc}  & 29.82 & 57.51 & 42.16 & 45.97 & 34.50 & 10.23 & 10.50 & 3.00 & 27.00 & 17.05 & 44.02 \\
\textbf{+DTR (Ours)}       & \best{52.63} & 46.63 & \best{77.45} & \best{55.99} & \best{52.00} & \best{28.41} & \best{29.50} & 10.50 & \best{38.50} & \best{31.78} & 44.12 \\

\bottomrule
\end{tabular}
}
\end{table*}

\begin{table*}[t]
\caption{Results on Qwen-VL-based video models, including Qwen2.5-VL-7B and Qwen3-VL-8B, across EventHallusion, VideoHallucer, and MVBench. Higher is better, and all values are reported as accuracy (\%). Bold indicates the best result.}
\label{tab:qwen_based_models}
\centering
\small
\renewcommand{\arraystretch}{1.08}
\resizebox{\textwidth}{!}{
\begin{tabular}{cccccccccccc}
\toprule
\multirow{2}{*}{\raisebox{-0.3ex}{Method}}
& \multicolumn{4}{c}{EventHallusion~\cite{zhang2024eventhallusion}}
& \multicolumn{6}{c}{VideoHallucer (Hallucinated)~\cite{wang2024videohallucer}}
& MVBench~\cite{li2024mvbench} \\
\cmidrule(lr){2-5}
\cmidrule(lr){6-11}
\cmidrule(lr){12-12}
& Entire & Mix & Misleading & Overall
& Object-Relation & Temporal & Semantic Detail & Factual & Non-factual & Overall
& Overall \\
\midrule

\rowcolor{ACMBlue!10}
\multicolumn{12}{c}{\textbf{Qwen2.5-VL-7B}~\cite{bai2025qwen25vltechnicalreport}} \\
\midrule
Baseline \cite{bai2025qwen25vltechnicalreport}    & 64.04 & 60.10 & 88.24 & 68.22 & 83.50 & 78.98 & 90.00 & 44.00 & 81.00 & 75.50 & 58.90 \\
+TCD \cite{zhang2024eventhallusion}        & 61.40 & 63.73 & 84.31 & 68.22 & 83.00 & 74.43 & 88.50 & 38.50 & 76.00 & 72.09 & 58.55 \\
+VCD \cite{leng2024mitigating}  & 59.65 & 60.10 & 86.27 & 66.50 & 80.50 & 73.86 & 86.50 & 44.00 & 80.50 & 73.07 & 57.20 \\
+DINO-HEAL \cite{li2025vidhalluc}  & \best{69.30} & 58.55 & 86.27 & 68.46 & 83.00 & 78.98 & 91.50 & 44.50 & 81.00 & 75.80 & 58.93  \\
\textbf{+DTR (Ours)}       & 65.79 & \best{69.95} & \best{90.20} & \best{73.84} & \best{84.50} & \best{81.25} & \best{92.00} & \best{49.00} & \best{82.00} & \best{77.75} & \best{59.25} \\

\midrule
\rowcolor{ACMBlue!10}
\multicolumn{12}{c}{\textbf{Qwen3-VL-8B}~\cite{bai2025qwen3}} \\
\midrule
Baseline \cite{bai2025qwen3}    & 62.28 & 65.28 & 96.08 & 72.13 & 80.00 & 73.86 & 81.00 & 52.50 & 91.50 & 75.77 & 66.00 \\
+TCD \cite{zhang2024eventhallusion}        & \best{63.16} & 65.80 & \best{97.06} & 72.86 & 80.00 & 73.30 & 80.50 & 54.00 & 90.50 & 75.66 & 66.67 \\
+VCD \cite{leng2024mitigating}  & 60.53 & 62.69 & 97.06 & 70.66 & 80.00 & 70.45 & 78.50 & 50.50 & 91.00 & 74.09 & 65.60 \\
+DINO-HEAL \cite{li2025vidhalluc}  & 62.28 & 66.84 & 96.08 & 72.86 & 76.00 & 74.43 & 76.50 & \best{61.50} & \best{93.50} & 76.39 & 66.17 \\
\textbf{+DTR (Ours)}       & \best{63.16} & \best{72.54} & 96.08 & \best{75.79} & \best{85.50} & \best{79.55} & \best{82.00} & 57.50 & 93.00 & \best{79.51} & \best{66.83} \\

\bottomrule
\end{tabular}
}
\end{table*}

\begin{figure*}[t]
    \captionsetup[subfigure]{labelfont=mysubcap}
    \centering

    \begin{subfigure}[t]{0.498\textwidth}
        \centering
        \includegraphics[width=\linewidth]{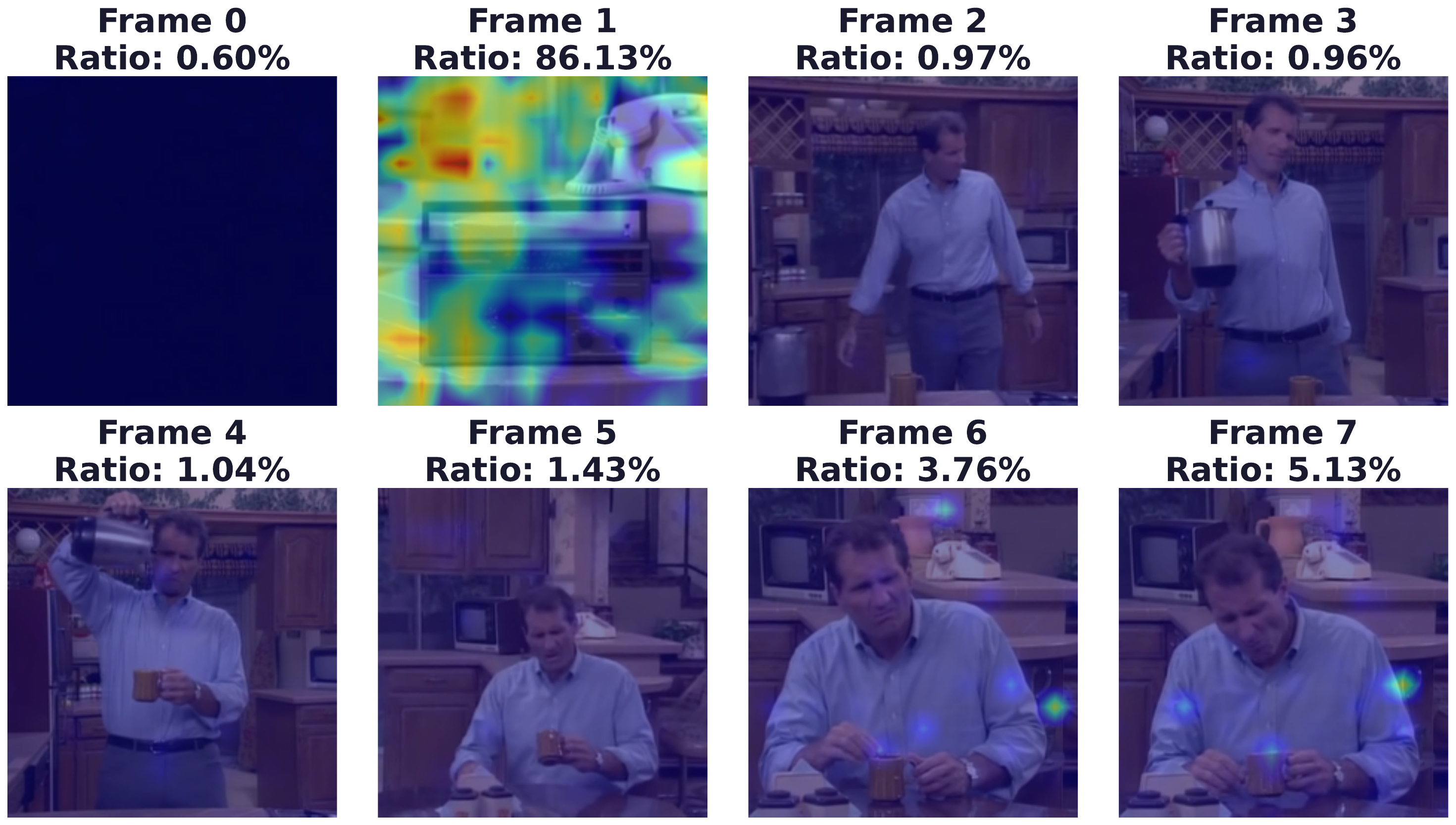}

        \vspace{0.0pt}
        {\centering
        \begin{minipage}{0.96\linewidth}
        \centering
        \fontsize{7}{11}\selectfont\fontfamily{qhv}\bfseries
        Q: Does 'he walks into the kitchen and pours a cup of coffee as he gloats' happen earlier than 'an old style radio is sitting on a counter'?\\
        A: Yes \qquad GT: No
        \end{minipage}
        \par}

        \caption{\fontsize{7}{11}\selectfont\fontfamily{qhv}\bfseries
        Before intervention. Averaged over decoder layers 18--20.}
        \label{fig:midlayer_attention_before}
    \end{subfigure}
    \begin{subfigure}[t]{0.498\textwidth}
        \centering
        \includegraphics[width=\linewidth]{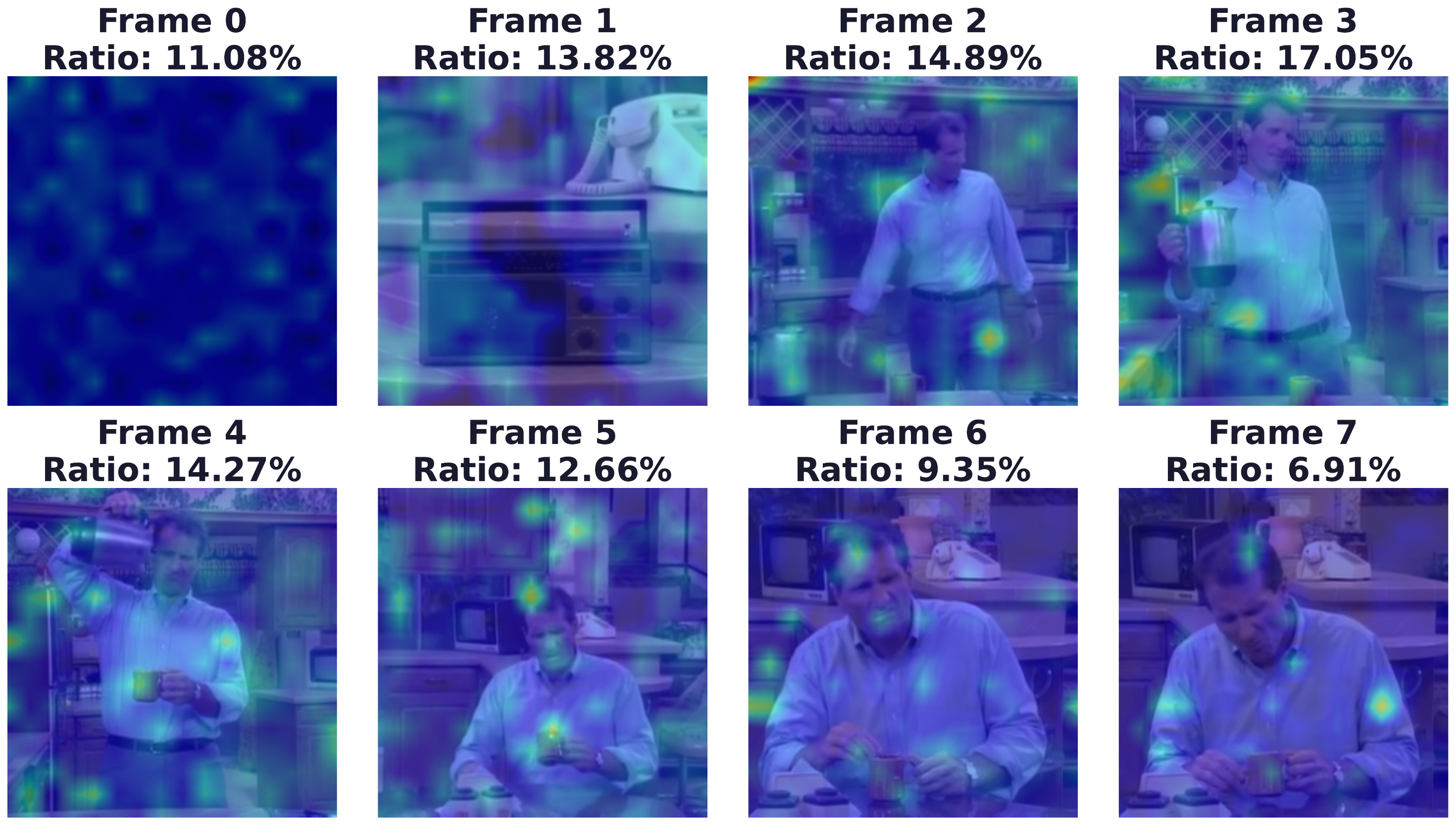}

        \vspace{0.0pt}
        {\centering
        \begin{minipage}{0.96\linewidth}
        \centering
        \fontsize{7}{11}\selectfont\fontfamily{qhv}\bfseries
        Q: Does 'he walks into the kitchen and pours a cup of coffee as he gloats' happen earlier than 'an old style radio is sitting on a counter'?\\
        A: No \qquad GT: No
        \end{minipage}
        \par}

        \caption{\fontsize{7}{11}\selectfont\fontfamily{qhv}\bfseries
        After intervention. Averaged over decoder layers 18--20.}
        \label{fig:midlayer_attention_after}
    \end{subfigure}

    \caption{Mid-layer attention before and after DTR on the same example. DTR redistributes attention across frames and corrects the answer from ``Yes'' to ``No.''}
    \label{fig:midlayer_attention_vis}
\end{figure*}

\section{Experiments}
\subsection{Experimental Settings}

\subsubsection{Datasets and Evaluation.}
We evaluate our method on two hallucination benchmarks, \textbf{EventHallusion} \cite{zhang2024eventhallusion} and \textbf{VideoHallucer} \cite{wang2024videohallucer}, and one general video understanding benchmark, \textbf{MVBench} \cite{li2024mvbench}. EventHallusion is used to assess event-level hallucinations caused by language priors and vision-language biases, and we follow its official binary QA setting and report accuracy.  VideoHallucer is an adversarial benchmark with paired \textit{basic} and \textit{hallucinated} questions for evaluating both intrinsic and extrinsic hallucinations in large video-language models. For LLaVA-based models, we follow its official protocol and report the paired \textit{Overall} accuracy. For Qwen-VL-based models, we report the accuracy on the \textit{hallucinated} questions as a diagnostic metric, as this subset more directly evaluates the model’s ability to reject hallucinated content in challenging settings. MVBench is adopted to measure general temporal video understanding ability across diverse tasks, and we report the average accuracy over all tasks.

\subsubsection{Models and Baselines.}
We evaluate our method on four representative open-source Video-LLMs in the main paper, including \textbf{LLaVA-NeXT-Video-7B}~\cite{zhang2024llavanextvideo}, \textbf{Video-LLaVA-7B}~\cite{lin2024video}, \textbf{Qwen2.5-VL-7B}~\cite{bai2025qwen25vltechnicalreport}, and \textbf{Qwen3-VL-8B}~\cite{bai2025qwen3}. These models cover diverse video-language architectures and temporal modeling behaviors. Additional validation on other Video-LLM variants is provided in Appendix~\ref{app:other_models}.
 As baselines, we consider three training-free hallucination mitigation methods: \textbf{TCD}~\cite{zhang2024eventhallusion}, a video-specific temporal contrastive decoding method that contrasts the outputs of the original video and a temporally degraded video to suppress event priors; \textbf{VCD}~\cite{leng2024mitigating}, which mitigates hallucinations by contrasting the outputs of the original input and a visually distorted counterpart to calibrate visual bias and language priors; and \textbf{DINO-HEAL}~\cite{li2025vidhalluc}, which reweights visual features using DINOv2-based saliency cues during inference.

\subsubsection{Implementation Details.}

Unless otherwise specified, we uniformly sample 8 frames from each input video and use greedy decoding for all models. In addition to the default 8-frame protocol, we further conduct auxiliary analyses under varied frame-sampling settings in Appendix~\ref{app:frame_sampling} to examine the robustness of the observed temporal concentration behavior and the effect of DTR beyond the main setup.
 Our method is a training-free, inference-time intervention that preserves the original attention computation implementation of each model and modifies decoder attention logits before softmax, while leaving the visual encoder and model parameters unchanged and introducing no extra decoding branches or auxiliary models. It is applied to visual-token attention in selected middle-to-late decoder layers to rebalance temporal evidence, alleviate anchor-frame over-dominance, and encourage broader multi-frame grounding during generation. We set the numerical stability constant in Eq.~\ref{eq:norm_gap} to \(\epsilon = 10^{-6}\). Unless otherwise noted, all methods are evaluated under the same input and decoding protocol. For baseline comparison, we follow the official implementations of VCD, TCD, and DINO-HEAL under their default inference settings, ensuring a fair, reproducible, and consistent comparison under the same evaluation protocol.

\begin{table*}[t]
\centering
\caption{Efficiency comparison on 100 samples from the \textit{entire} subset of EventHallusion. Each method is run three times, and the reported statistics are computed over all 300 inferences. Lower is better for time and memory, while higher is better for throughput.}
\label{tab:efficiency_eventhallusion_entire}
\small
\setlength{\tabcolsep}{5pt}
\resizebox{\textwidth}{!}{
\begin{tabular}{lccccccc}
\toprule
Method & Samples & Avg. Time (s)$\downarrow$ & Std. (s)$\downarrow$ & Min. (s)$\downarrow$ & Max. (s)$\downarrow$ & Mean Memory (MB)$\downarrow$ & Throughput (samples/s)$\uparrow$ \\
\midrule
Baseline \cite{lin2024video}   & 300 & 0.607 & 0.047 & 0.485 & 0.728 & 16697.156 & 1.647 \\
+TCD \cite{zhang2024eventhallusion}        & 300 & 1.187 & 0.060 & 0.976 & 1.321 & 18565.375 & 0.843 \\
+VCD \cite{leng2024mitigating}        & 300 & 3.437 & 0.702 & 1.603 & 3.968 & 19195.997 & 0.291 \\
+DINO-HEAL \cite{li2025vidhalluc}   & 300 & 0.785 & 0.209 & \best{0.535} & 1.229 & 17277.869 & 1.275 \\
\textbf{+DTR (Ours)} & 300 & \best{0.645} & \best{0.046} & 0.538 & \best{0.751} & \best{16697.163} & \best{1.550} \\
\bottomrule
\end{tabular}
}
\end{table*}

\subsection{Main Results}

\subsubsection{Results on LLaVA-based video models.}
Table~\ref{tab:llava_based_models} reports the results on two LLaVA-based Video-LLMs, i.e., LLaVA-NeXT-Video-7B and Video-LLaVA-7B. Our method consistently improves hallucination robustness on EventHallusion and VideoHallucer, while maintaining comparable performance on MVBench. On EventHallusion, the gains are more pronounced on the \textit{Misleading} and \textit{Entire} subsets, suggesting improved mitigation of language-prior bias and anchor-frame over-reliance. 
The improvement pattern also varies across subsets and model families, reflecting differences in benchmark characteristics and model behavior.

\subsubsection{Results on Qwen-VL-based models.}
Table~\ref{tab:qwen_based_models} summarizes the results on Qwen2.5-VL-7B and Qwen3-VL-8B. Our method consistently outperforms vanilla decoding, with EventHallusion overall improvements of 5.62 and 3.66 points, respectively, and remains competitive with other training-free baselines, demonstrating good generality across Qwen-VL-based backbones and strong cross-model robustness. On VideoHallucer, DTR yields consistent gains across multiple categories, especially temporal settings, while also improving factual settings. This suggests that DTR mainly strengthens temporal grounding and, more broadly, improves video grounding during generation. On MVBench, we report only the overall score and observe that DTR improves hallucination robustness without hurting video understanding.

\subsubsection{Efficiency Comparison on the EventHallusion Entire Subset}
To further evaluate practical inference cost, we compare different methods on 100 samples from the \textit{Entire} subset of EventHallusion. Each method is run three times, and we report aggregated statistics over all 300 inferences, including average latency, standard deviation, minimum/maximum latency, mean memory usage, and throughput. As shown in Table~\ref{tab:efficiency_eventhallusion_entire}, the baseline remains the most efficient. DTR (Ours) achieves a better trade-off between efficiency and effectiveness than other mitigation methods, while remaining close to the baseline. In contrast, VCD incurs the highest additional inference cost.

\subsection{Attention Analysis}
We analyze how DTR affects temporal evidence aggregation from both qualitative and quantitative perspectives. Qualitatively, Figure~\ref{fig:midlayer_attention_vis} shows that the model exhibits severe anchor-frame bias before intervention, whereas DTR redistributes attention more evenly across frames and corrects the prediction. Quantitatively, Table~\ref{tab:dtr_mechanism} shows that global-only rebalance ($\alpha=0.5,\beta=0$) already reduces temporal imbalance, lowering the dominance ratio from 0.746 to 0.558 and increasing entropy and non-anchor frame ratio from 0.990/0.254 to 1.495/0.442. A compensation-only variant ($\alpha=0,\beta=0.3$) further improves these attention statistics, with dominance ratio, entropy, and non-anchor frame ratio reaching 0.490, 1.656, and 0.510, respectively, but achieves only 28.59 average accuracy. DTR ($\alpha=0.5,\beta=0.3$) performs best overall, reducing the dominance ratio to 0.195, increasing entropy and non-anchor frame ratio to 2.027 and 0.863, and achieving the highest average accuracy of 43.22. These results show that while either component alone can alleviate temporal imbalance, combining them yields the most effective redistribution of temporal evidence and the strongest hallucination mitigation.

\begin{table}[t]
\caption{Attention analysis on Video-LLaVA-7B over the Object-Relation and Temporal subsets from VideoHallucer. Dom. denotes the average anchor-frame dominance ratio, Entropy measures the dispersion of frame-level attention, and Non-anchor denotes the total attention mass assigned to non-anchor frames.}
\label{tab:dtr_mechanism}
\centering
\small
\setlength{\tabcolsep}{4pt}
\begin{tabular*}{\columnwidth}{@{\extracolsep{\fill}}lcccc@{}}
\toprule
Setting & Dom.$\downarrow$ & Ent.$\uparrow$ & Non-anchor$\uparrow$ & Avg. Acc.$\uparrow$ \\
\midrule
Baseline & 0.746 & 0.990 & 0.254 & 21.26 \\
Global-only & 0.558 & 1.495 & 0.442 & 37.59 \\
Comp.-only  & 0.490 & 1.656 & 0.510 & 28.59 \\
DTR  & \best{0.195} & \best{2.027} & \best{0.863} & \best{43.22} \\
\bottomrule
\end{tabular*}
\end{table}

\begin{figure*}[t]
    \centering
    \includegraphics[width=\linewidth]{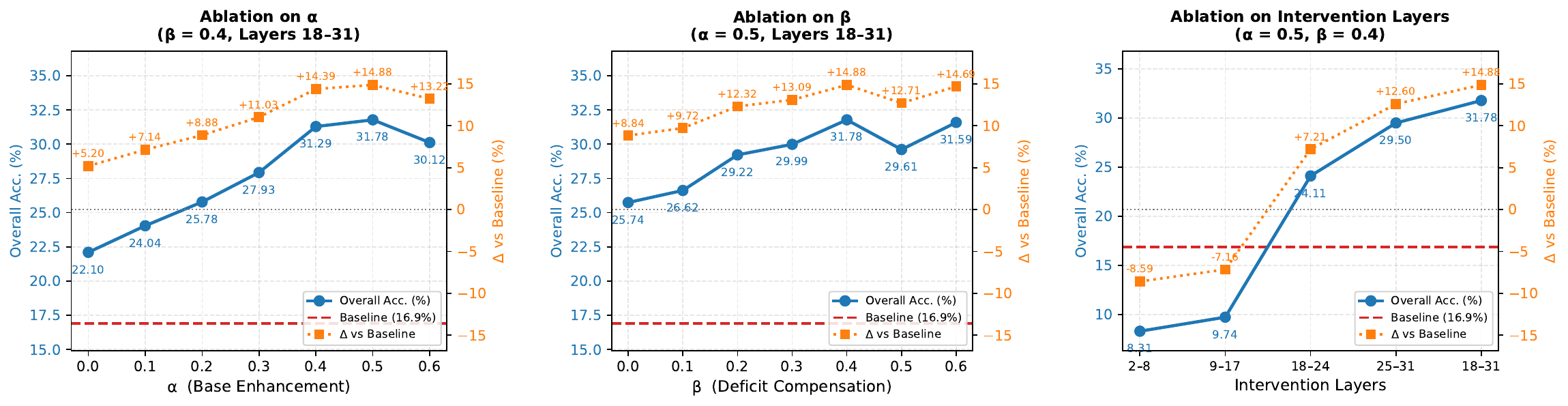}
    \caption{
        Ablation study on Video-LLaVA-7B (VideoHallucer).
        Each plot shows Overall Accuracy (\%) and gain $\Delta$ over the baseline
        on the left and right y-axes, respectively.
        \textbf{Left:} Effect of global adjustment $\alpha$ ($\beta{=}0.4$, layers 18--31).
        \textbf{Middle:} Effect of deficit compensation $\beta$ ($\alpha{=}0.5$, layers 18--31).
        \textbf{Right:} Effect of intervention layer range ($\alpha{=}0.5$, $\beta{=}0.4$).
        The dashed line marks the baseline accuracy (16.9\%).
    }
    \label{fig:ablation}
\end{figure*}
\subsection{Ablation Study}

We conduct all ablation experiments on \textbf{Video-LLaVA-7B} and report results on \textbf{VideoHallucer}. Unless otherwise specified, the default setting uses $\alpha=0.5$, $\beta=0.4$, and decoder layers 18--31. We report the official \textbf{Overall Accuracy (\%)} on VideoHallucer.

\subsubsection{Effect of Shared Adjustment Strength $\alpha$.}
We vary $\alpha$ while fixing $\beta=0.4$ and the intervention layers to 18--31. As shown in Figure~\ref{fig:ablation} (Left), performance improves steadily as $\alpha$ increases from 0.0 to 0.5 and peaks at $\alpha=0.5$. This trend indicates that an appropriate shared adjustment is important for the effectiveness of DTR, since it provides a stable basis for subsequent frame-relative reallocation under anchor-frame dominance. When $\alpha$ is too small, the redistribution remains insufficient and non-anchor frames are still weakly utilized. With a moderate $\alpha$, DTR can more effectively relax anchor over-dominance and promote a broader use of temporal evidence across frames. Further increasing $\alpha$ slightly reduces the gain, suggesting that an overly strong shared adjustment may weaken the relative selectivity needed for precise temporal grounding.

\subsubsection{Effect of Deficit Compensation $\beta$.}
We then vary $\beta$ while fixing $\alpha=0.5$ and the intervention layers to 18--31. As shown in Figure~\ref{fig:ablation} (middle), the case of $\beta=0$ reduces DTR to a shared-adjustment-only control, in which all frames receive the same adjustment from $\alpha$ and no frame-relative compensation is applied. This variant already outperforms the baseline, suggesting that the shared adjustment alone can partially alleviate temporal imbalance. However, introducing $\beta$ yields further gains and achieves the best result at $\beta=0.4$, indicating that shared adjustment alone is insufficient and that adaptive compensation for under-attended frames plays a critical role in correcting temporal imbalance. This supports our motivation that mitigating anchor-frame dominance requires both a shared adjustment and targeted rebalancing of weaker frames. When $\beta$ becomes too large, the gains become less stable, suggesting that excessive compensation may weaken the selectivity of temporal evidence allocation and introduce unnecessary disturbance to the decoding process.

\subsubsection{Effect of Intervention Layers.}
Finally, we study the effect of the intervention layer range by fixing $\alpha=0.5$ and $\beta=0.4$. As shown in Figure~\ref{fig:ablation} (Right), intervening in early layers hurts performance, whereas middle-to-late and upper-layer intervention brings clear gains, with layers 18--31 performing best. We attribute this pattern to the stage at which anchor-frame dominance begins to affect temporal evidence aggregation. Although anchor-frame bias is already visible in early decoder layers, it has not yet become the main bottleneck for generation. Figure~\ref{fig:visual_ratio} further shows that visual attention remains suppressed over a sustained middle-to-late range, suggesting that decoding is increasingly driven by a narrow temporal evidence path. Therefore, intervention in middle-to-late layers is more effective, as it relaxes anchor-frame over-dominance before the imbalance is reflected in the final output. This is also consistent with Figure~\ref{fig:midlayer_attention_vis}, where DTR redistributes attention from the anchor frame to a broader set of frames and corrects the answer. Overall, the most effective intervention region lies in the middle-to-late decoder layers.

\section{Conclusion and Future Work}
In this paper, we identify anchor-frame dominance as a key source of hallucination in Video-LLMs, where decoding tends to over-focus on a nearly fixed temporal position and consequently underuses broader temporal evidence. To address this issue, we propose DTR, a simple training-free and layer-selective decoding strategy that rebalances decoder-side visual attention without modifying the visual encoder or requiring additional training. Extensive experiments on both hallucination and general video understanding benchmarks across multiple backbones show that DTR consistently improves hallucination robustness while preserving strong video understanding performance and high inference efficiency, suggesting that balanced temporal evidence allocation is an important principle for faithful video understanding. As future work, it would be valuable to study the role of temporal faithfulness in broader real-world and safety-critical video understanding scenarios.

\bibliographystyle{ACM-Reference-Format}
\bibliography{sample-base}
\appendix
\FloatBarrier
\begin{figure}[!b]
    \centering
    \includegraphics[width=1\columnwidth]{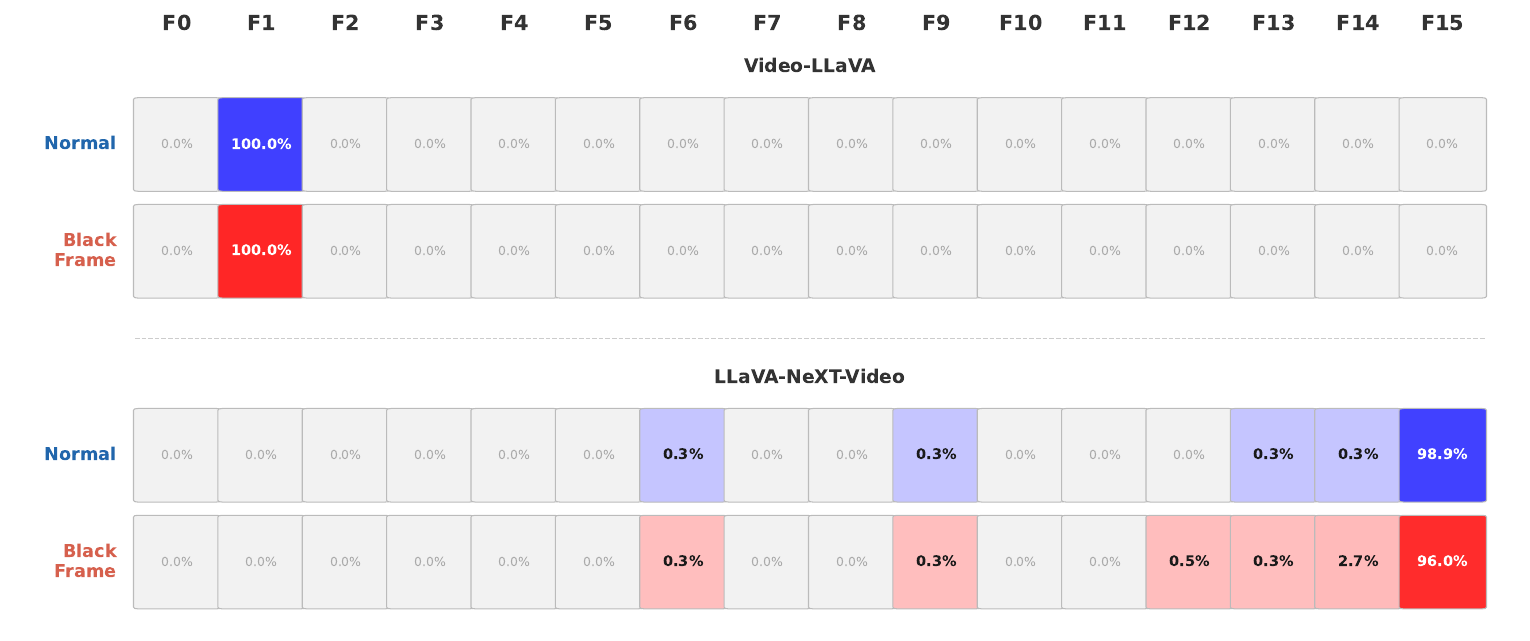}
    \caption{Anchor-position statistics under normal inputs and black-frame intervention for Video-LLaVA-7B and LLaVA-NeXT-Video-7B under 16-frame sampling. In the black-frame setting, the anchor frame identified from the normal input is replaced with an all-black frame before visual encoding. Percentages denote how often each frame is selected as the anchor frame on the Object-Relation and Temporal subsets of VideoHallucer.}

    \label{fig:anchor_bias_16frames}
\end{figure}
\begin{figure}[!h]
    \captionsetup[subfigure]{labelfont=mysubcap}
    \centering

    \begin{subfigure}[t]{0.48\columnwidth}
        \centering
        \includegraphics[width=\linewidth]{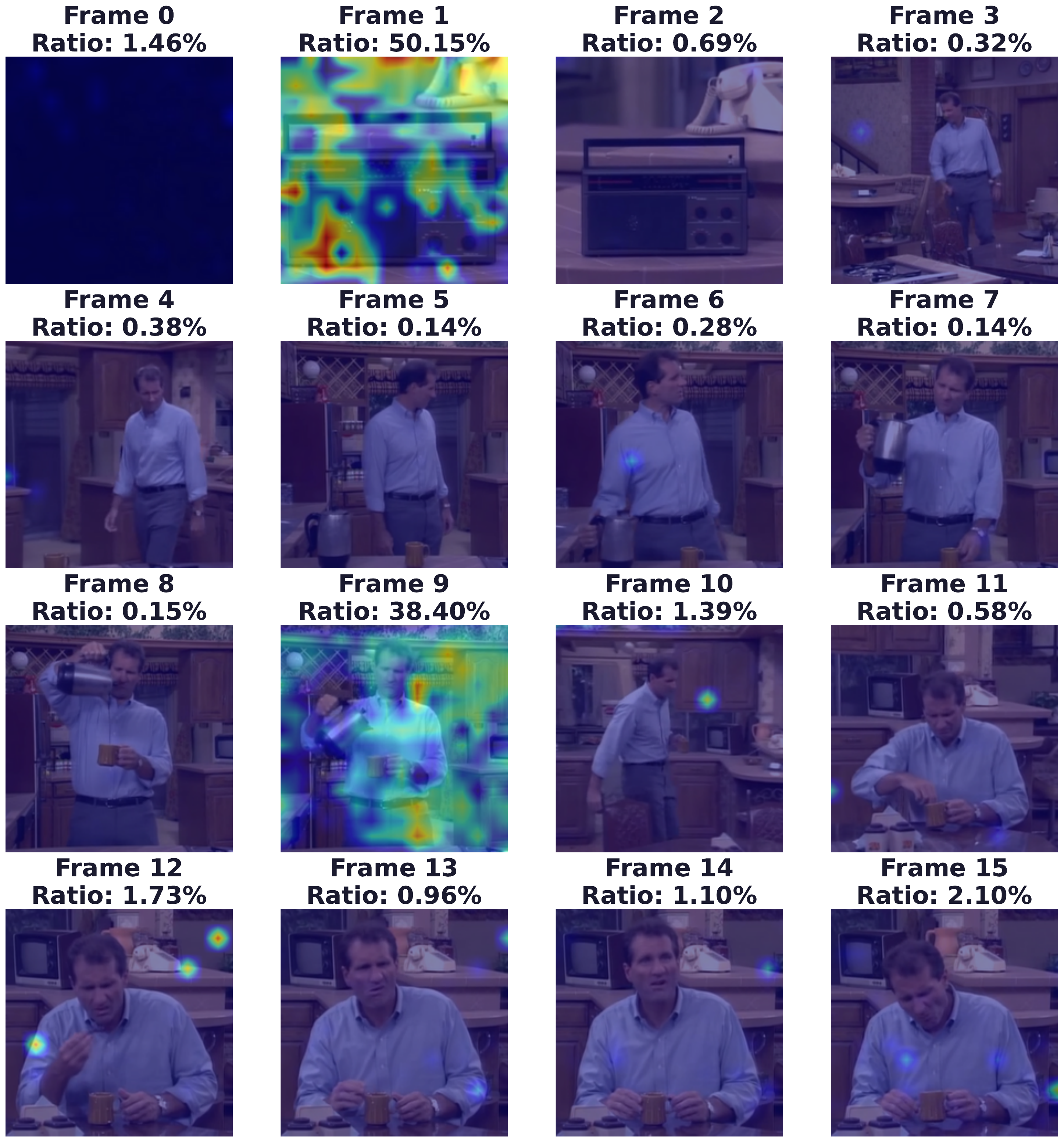}
        \caption{Video-LLaVA}
        \label{fig:videollava_16_baseline}
    \end{subfigure}\hfill
    \begin{subfigure}[t]{0.48\columnwidth}
        \centering
        \includegraphics[width=\linewidth]{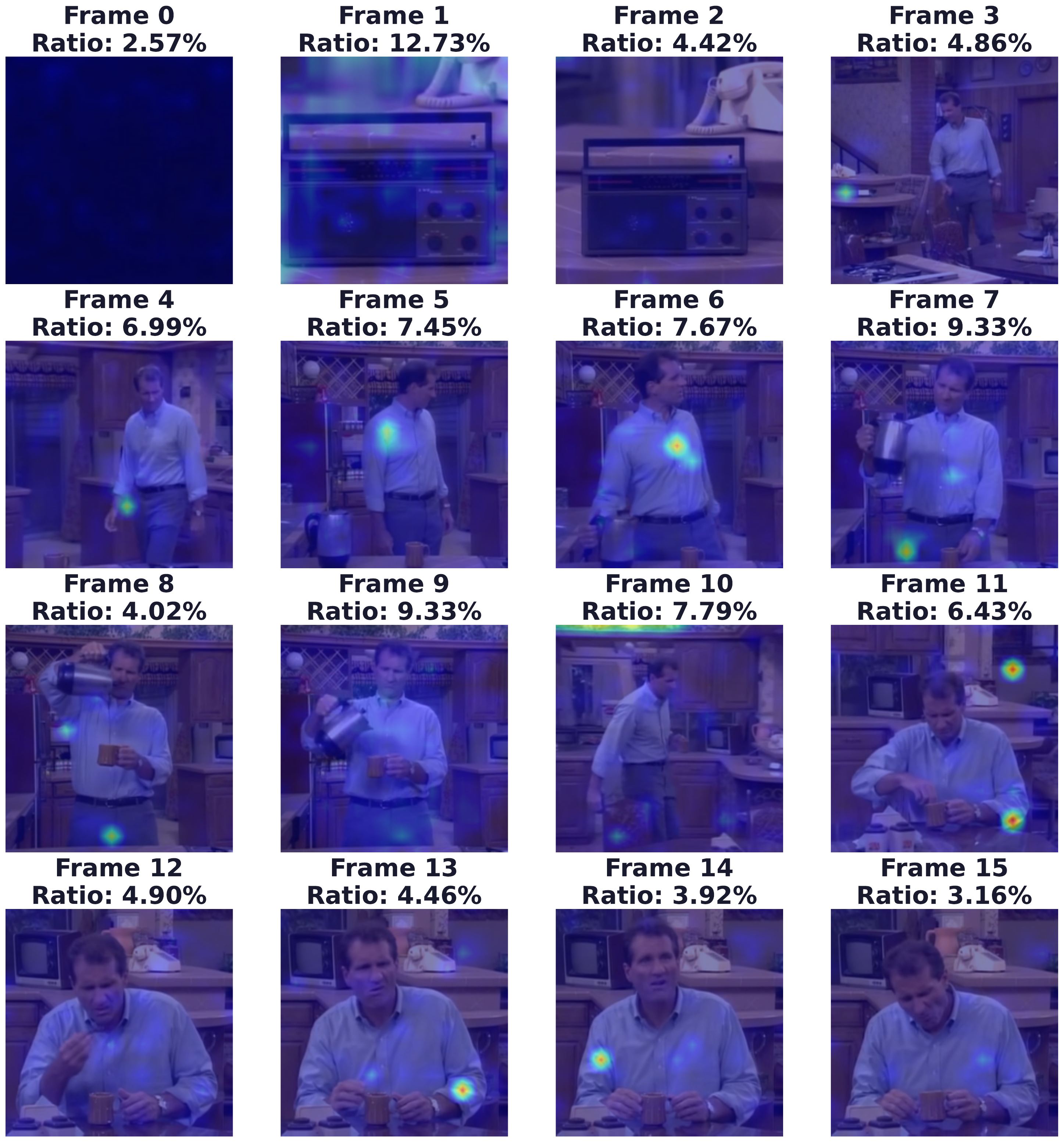}
        \caption{Video-LLaVA (+DTR)}
        \label{fig:videollava_16_dtr}
    \end{subfigure}

    \vspace{0.3em}

    \begin{subfigure}[t]{0.48\columnwidth}
        \centering
        \includegraphics[width=\linewidth]{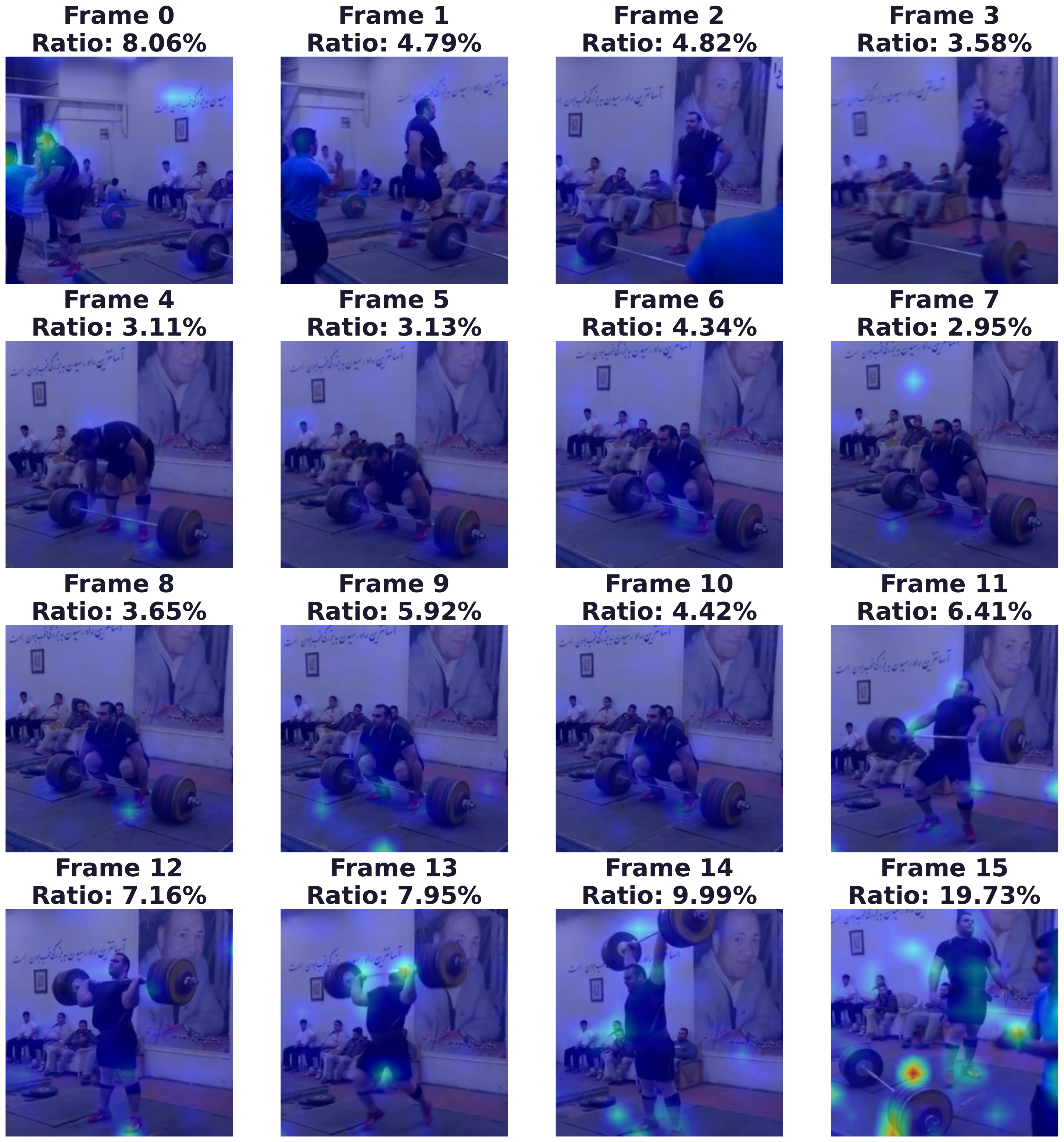}
        \caption{LLaVA-NeXT-Video}
        \label{fig:llavanext_16_baseline}
    \end{subfigure}\hfill
    \begin{subfigure}[t]{0.48\columnwidth}
        \centering
        \includegraphics[width=\linewidth]{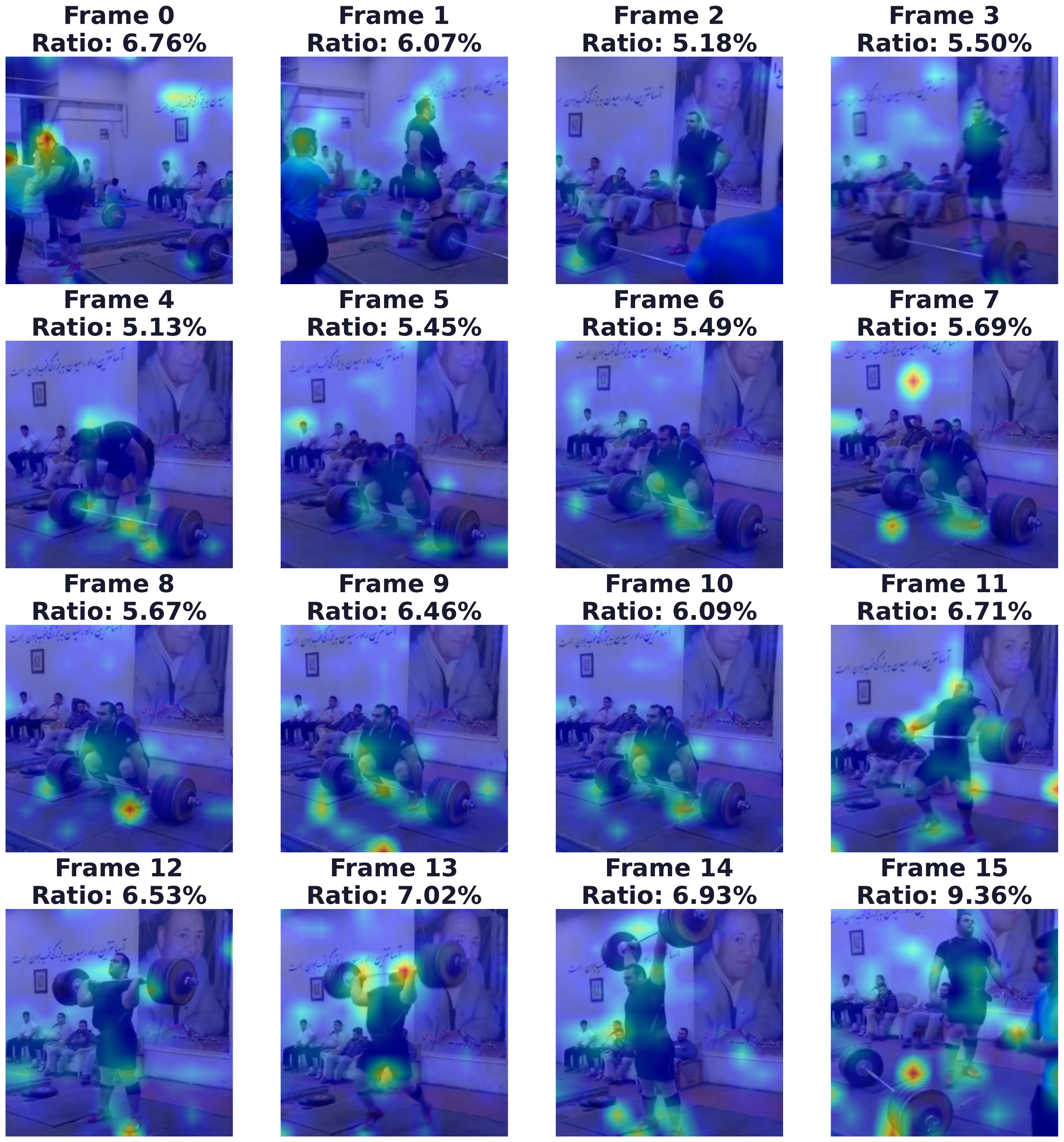}
        \caption{LLaVA-NeXT-Video (+DTR)}
        \label{fig:llavanext_16_dtr}
    \end{subfigure}

    \caption{Attention visualizations under 16-frame sampling for Video-LLaVA-7B and LLaVA-NeXT-Video-7B.}
    \label{fig:attention_visualizations_16frame}
\end{figure}
\section{Robustness Across Frame Sampling}
\label{app:frame_sampling}
We further examine whether the findings presented earlier in the paper persist under different frame-sampling densities.

Figure~\ref{fig:anchor_bias_16frames} first shows that, for Video-LLaVA-7B~\cite{lin2024video} and LLaVA-NeXT-Video-7B~\cite{zhang2024llavanextvideo}, the black-frame finding reported earlier in the paper still holds under 16-frame sampling: after replacing the anchor frame with an all-black frame before visual encoding, the dominant anchor position remains nearly unchanged. This again suggests that, in these models, the dominant position mainly reflects a persistent structural or positional bias rather than adaptive content-dependent grounding. Consistent with this, the 16-frame visualizations in Figure~\ref{fig:attention_visualizations_16frame} shows that the baseline still concentrates heavily on a small subset of frames, whereas \method{} redistributes attention more broadly across frames, reflecting its decoder-side mechanism for amplifying the grounding contribution of under-attended frames during generation.

We next summarize the quantitative results by model family. For Qwen2.5-VL-7B~\cite{bai2025qwen25vltechnicalreport}, Table~\ref{tab:qwen25vl_frame_comparison} shows that \method{} consistently improves over the baseline across all tested frame-sampling settings, from 4 to 20 frames, suggesting that its benefit is not tied to a single temporal resolution but remains stable across different sampling densities. For LLaVA-NeXT-Video-7B, we further report results under the denser 16-frame setting to examine the effect of higher frame density.
 As shown in Table~\ref{tab:llavanextvideo_16frame_results}, \method{} still improves performance on EventHallusion~\cite{zhang2024eventhallusion}, VideoHallucer~\cite{wang2024videohallucer}, and MVBench~\cite{li2024mvbench}.

\begin{table}[!b]
\caption{Comparison of baseline and \textbf{+DTR} on Qwen2.5-VL-7B under different frame sampling settings on \textbf{EventHallusion}. We report only the \textit{All} accuracy (\%). Higher is better.}
\label{tab:qwen25vl_frame_comparison}
\centering
\small
\setlength{\tabcolsep}{8pt}
\begin{tabular}{cccc}
\toprule
Frames & Baseline & \textbf{+DTR (Ours)} & $\Delta$ \\
\midrule
4  & 65.53 & \textbf{71.39} & +5.86 \\
6  & 68.46 & \textbf{73.11} & +4.65 \\
8  & 68.22 & \textbf{73.84} & +5.62 \\
16 & 68.70 & \textbf{73.84} & +5.14 \\
18 & 66.99 & \textbf{73.59} & +6.60 \\
20 & 67.73 & \textbf{74.57} & +6.84 \\
\bottomrule
\end{tabular}
\end{table}

\begin{table*}[!t]
\caption{Comparison between the baseline and \textbf{+DTR} on LLaVA-NeXT-Video-7B under the 16-frame setting across EventHallusion, VideoHallucer, and MVBench. Higher is better, and all values are reported as accuracy (\%). }
\label{tab:llavanextvideo_16frame_results}
\centering
\small
\setlength{\tabcolsep}{4.4pt}
\renewcommand{\arraystretch}{1.06}
\begin{adjustbox}{width=\textwidth,center}
\begin{tabular}{lccccccccccc}
\toprule
\multirow{2}{*}{\textbf{Method}}
& \multicolumn{4}{c}{\textbf{EventHallusion}~\cite{zhang2024eventhallusion}}
& \multicolumn{6}{c}{\textbf{VideoHallucer}~\cite{wang2024videohallucer}}
& \textbf{MVBench}~\cite{li2024mvbench} \\
\cmidrule(lr){2-5}
\cmidrule(lr){6-11}
\cmidrule(lr){12-12}
& Entire & Mix & Misleading & Overall
& Object-Relation & Temporal & Semantic Detail & Factual & Non-factual & Overall
& Overall \\
\midrule
LLaVA-NeXT-Video-7B~\cite{zhang2024llavanextvideo} & 46.49 & \best{68.91} & 59.80 & 60.39 & 50.00 & 11.93 & 42.00 & 4.00 & 32.00 & 27.99 & 43.93 \\
\textbf{+DTR (Ours)} & \textbf{54.39} & 62.18 & \textbf{70.59} & \textbf{62.10} & \textbf{52.50} & \textbf{20.45} & \textbf{47.00} & \textbf{7.50} & \textbf{37.50} & \textbf{32.99} & \textbf{44.53} \\
\bottomrule
\end{tabular}
\end{adjustbox}
\end{table*}
\begin{figure}[!t]
    \centering
    \includegraphics[width=1\columnwidth]{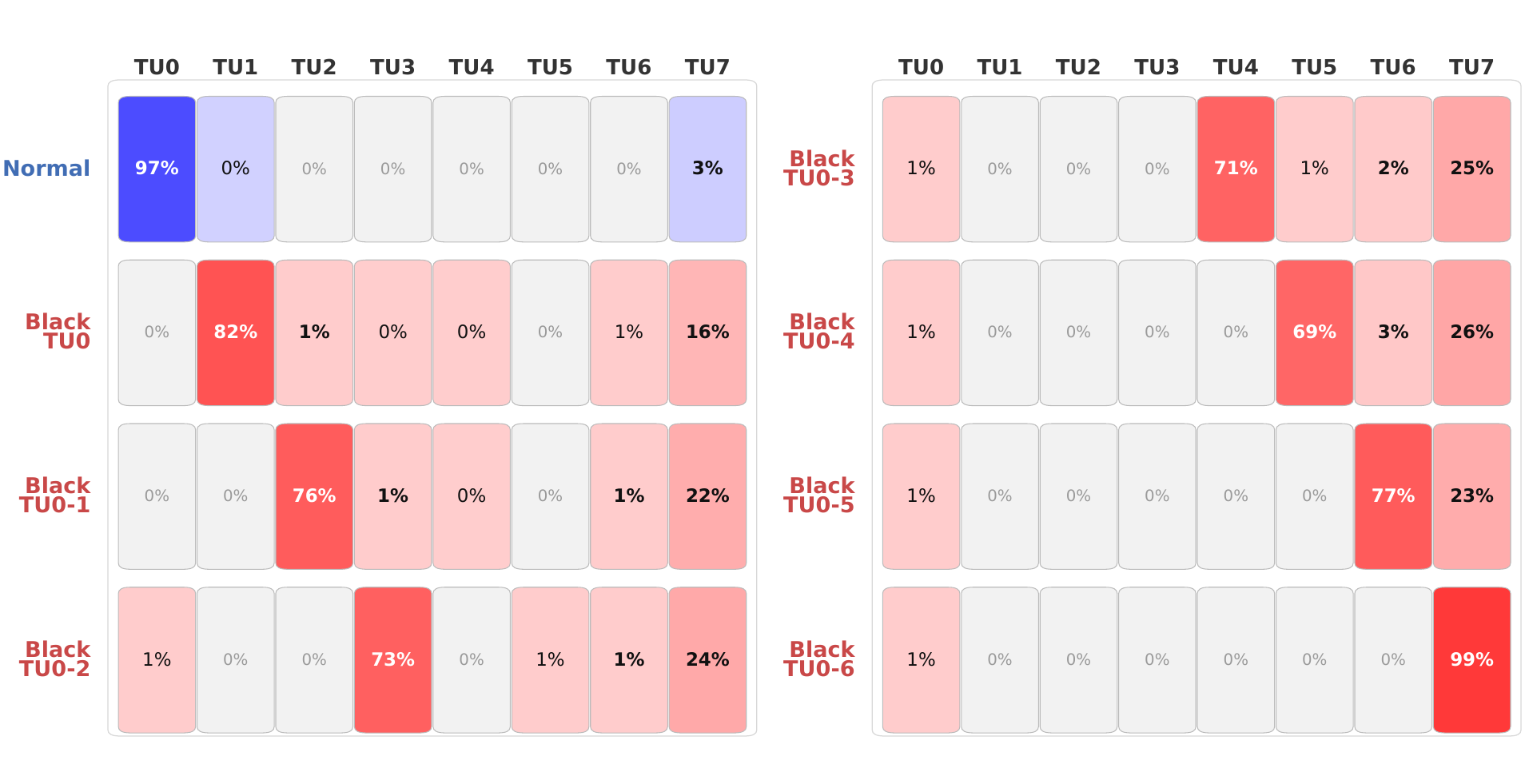}
    \caption{Black-frame analysis on Qwen2.5-VL-7B, aggregated over the \textbf{Object-Relation} and \textbf{Temporal} subsets of \textbf{VideoHallucer}. Since Qwen2.5-VL-7B groups every two consecutive frames into one temporal unit, we perform black-frame intervention at the temporal-unit level by simultaneously replacing the two frames within a unit with black frames. Starting from the normal input, we progressively mask temporal units from the beginning, i.e., \textit{Black TU0}, \textit{Black TU0-1}, \ldots, \textit{Black TU0-6}. Percentages denote how often each temporal unit is selected as the anchor.}
\label{fig:qwen_blackframe}
\end{figure}

\begin{figure}[!t]
    \centering
    \includegraphics[width=1\columnwidth]{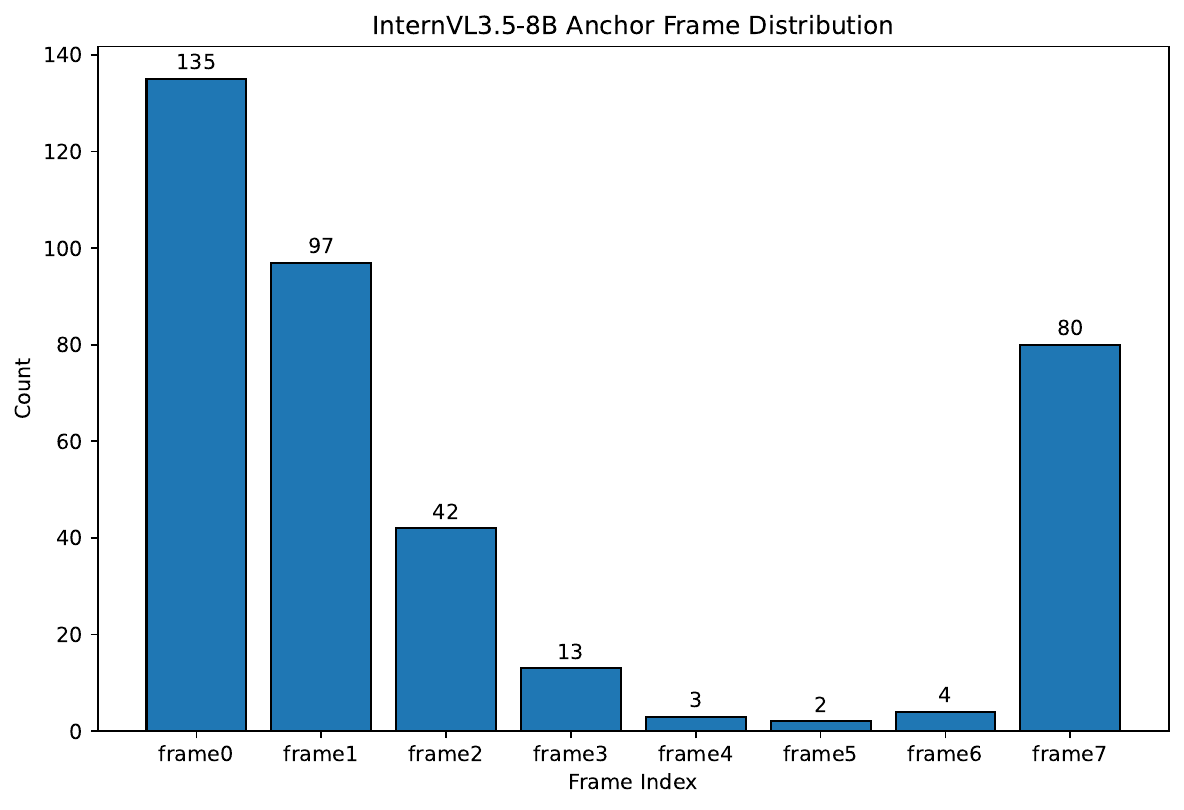}
    \caption{Anchor-frame distribution of InternVL3.5-8B, aggregated over the \textbf{Object-Relation} and \textbf{Temporal} subsets of \textbf{VideoHallucer}. The count at each frame position indicates how many times that frame is selected as the anchor across the evaluated samples. The model shows a clear overall preference for positions near the temporal boundaries.}
    \label{fig:internvl_anchor_dist}
\end{figure}
\section{Additional Anchor-Frame Statistics}
\label{app:add_anchor_stats}

We further examine whether the anchor-selection behavior observed earlier in the paper also appears beyond the LLaVA family. For Qwen2.5-VL-7B~\cite{bai2025qwen25vltechnicalreport}, the black-frame analysis is conducted at the level of temporal units rather than individual frames, since the model groups every two consecutive frames together when processing video inputs. Accordingly, in our intervention, the two frames within the same temporal unit are simultaneously replaced with black frames. Figure~\ref{fig:qwen_blackframe} reports the resulting anchor-selection pattern across different masking settings. The model tends to assign anchors to the first meaningful temporal unit and to the last temporal unit. InternVL3.5-8B~\cite{wang2025internvl3} exhibits a similar non-uniform pattern in Figure~\ref{fig:internvl_anchor_dist}, where anchor frames are concentrated near the beginning and the end of the sequence while middle frames are selected much less frequently.

Taken together with the black-frame results reported earlier in the paper, these observations suggest that temporal boundary preference recurs across multiple Video-LLM backbones, although its exact form varies by architecture. Overall, both the earlier results in the paper and the results in this appendix remain consistent with the interpretation that anchor-frame dominance mainly reflects a persistent structural or positional bias rather than fully content-adaptive grounding.

\section{Additional Results on Other Models}
\label{app:other_models}

\begin{table*}[!t]
\caption{Results on Qwen3-VL-4B across EventHallusion, VideoHallucer, and MVBench. Higher is better, and all values are reported as accuracy (\%). Bold indicates the best result in each column.}
\label{tab:qwen3vl_4b_results}
\centering
\small
\setlength{\tabcolsep}{4.2pt}
\renewcommand{\arraystretch}{1.06}
\begin{adjustbox}{width=\textwidth,center}
\begin{tabular}{lccccccccccc}
\toprule
\multirow{2}{*}{\textbf{Method}}
& \multicolumn{4}{c}{\textbf{EventHallusion}~\cite{zhang2024eventhallusion}}
& \multicolumn{6}{c}{\textbf{VideoHallucer (Hallucinated)}~\cite{wang2024videohallucer}}
& \textbf{MVBench}~\cite{li2024mvbench} \\
\cmidrule(lr){2-5}
\cmidrule(lr){6-11}
\cmidrule(lr){12-12}
& Entire & Mix & Misleading & Overall
& Object-Relation & Temporal & Semantic Detail & Factual & Non-factual & Overall
& Overall \\
\midrule
Qwen3-VL-4B~\cite{bai2025qwen3} & 65.79 & 49.74 & 97.06 & 66.01 & 88.00 & 82.39 & \best{89.00} & 53.00 & 86.00 & 79.68 & 64.48 \\
+TCD~\cite{zhang2024eventhallusion} & 62.28 & 51.81 & 97.06 & 66.01 & 85.50 & 81.82 & 88.00 & 52.50 & 86.50 & 78.86 & 63.39 \\
+VCD~\cite{leng2024mitigating} & 65.79 & 49.22 & \best{98.04} & 66.01 & 85.00 & 76.70 & 84.00 & 50.00 & 84.50 & 76.04 & \best{67.60} \\
+DINO-HEAL~\cite{li2025vidhalluc} & 64.04 & 47.67 & 97.06 & 64.55 & 87.50 & 82.39 & \best{89.00} & 51.50 & 87.50 & 79.58 & 64.50 \\
\textbf{+DTR (Ours)} & \best{68.42} & \best{59.07} & 97.06 & \best{71.15} & \best{89.50} & \best{85.80} & \best{89.00} & \best{56.00} & \best{91.50} & \best{82.36} & 64.58 \\
\bottomrule
\end{tabular}
\end{adjustbox}
\end{table*}

\begin{table*}[!t]
\caption{Results on InternVL3.5-8B across EventHallusion, VideoHallucer, and MVBench. Higher is better, and all values are reported as accuracy (\%). Bold indicates the best result in each column.}
\label{tab:internvl35_8b_results}
\centering
\small
\setlength{\tabcolsep}{4.2pt}
\renewcommand{\arraystretch}{1.06}
\begin{adjustbox}{width=\textwidth,center}
\begin{tabular}{lccccccccccc}
\toprule
\multirow{2}{*}{\textbf{Method}}
& \multicolumn{4}{c}{\textbf{EventHallusion}~\cite{zhang2024eventhallusion}}
& \multicolumn{6}{c}{\textbf{VideoHallucer}~\cite{wang2024videohallucer}}
& \textbf{MVBench}~\cite{li2024mvbench} \\
\cmidrule(lr){2-5}
\cmidrule(lr){6-11}
\cmidrule(lr){12-12}
& Entire & Mix & Misleading & Overall
& Object-Relation & Temporal & Semantic Detail & Factual & Non-factual & Overall
& Overall \\
\midrule
InternVL3.5-8B~\cite{wang2025internvl3} & 54.39 & 35.75 & \best{87.25} & 53.79 & 42.00 & \best{69.89} & 58.50 & 21.50 & 53.00 & 48.98 & 70.62 \\
+TCD~\cite{zhang2024eventhallusion} & 57.89 & \best{42.49} & 86.27 & 57.70 & \best{46.50} & 69.32 & \best{60.00} & 20.50 & 52.50 & 49.76 & 69.42 \\
+VCD~\cite{leng2024mitigating} & 58.77 & 39.90 & 84.31 & 56.23 & 38.00 & 54.55 & 46.50 & \best{24.50} & 48.50 & 42.41 & 66.15 \\
+DINO-HEAL~\cite{li2025vidhalluc} & 54.39 & 37.82 & \best{87.25} & 54.77 & 43.50 & 68.75 & 58.50 & 21.00 & 53.50 & 49.05 & 70.38 \\
\textbf{+DTR (Ours)} & \best{62.28} & 41.97 & \best{87.25} & \best{58.92} & 44.50 & \best{69.89} & 57.50 & 24.00 & \best{54.50} & \best{50.08} & \best{70.73} \\
\bottomrule
\end{tabular}
\end{adjustbox}
\end{table*}

\begin{table*}[!t]
\caption{Results on InternVL3.5-14B across EventHallusion, VideoHallucer, and MVBench. Higher is better, and all values are reported as accuracy (\%). Bold indicates the best result in each column.}
\label{tab:internvl35_14b_results}
\centering
\small
\setlength{\tabcolsep}{4.2pt}
\renewcommand{\arraystretch}{1.06}
\begin{adjustbox}{width=\textwidth,center}
\begin{tabular}{lccccccccccc}
\toprule
\multirow{2}{*}{\textbf{Method}}
& \multicolumn{4}{c}{\textbf{EventHallusion}~\cite{zhang2024eventhallusion}}
& \multicolumn{6}{c}{\textbf{VideoHallucer}~\cite{wang2024videohallucer}}
& \textbf{MVBench}~\cite{li2024mvbench} \\
\cmidrule(lr){2-5}
\cmidrule(lr){6-11}
\cmidrule(lr){12-12}
& Entire & Mix & Misleading & Overall
& Object-Relation & Temporal & Semantic Detail & Factual & Non-factual & Overall
& Overall \\
\midrule
InternVL3.5-14B~\cite{wang2025internvl3}
& 40.35 & 47.67 & \best{82.35} & 54.28
& 50.00 & 67.61 & 55.00 & 9.50 & 43.50 & 45.12
& 70.38 \\
+TCD~\cite{zhang2024eventhallusion}
& 41.23 & 50.26 & 79.41 & 55.01
& \best{51.00} & 64.77 & 55.50 & \best{14.00} & \best{48.00} & 46.65
& 69.50 \\
+VCD~\cite{leng2024mitigating}
& 37.72 & 45.08 & 77.45 & 51.10
& 49.50 & 62.50 & \best{56.00} & 12.00 & 44.50 & 44.90
& \best{71.20} \\
+DINO-HEAL~\cite{li2025vidhalluc}
& 40.35 & 47.15 & \best{82.35} & 54.03
& 50.00 & 67.05 & 55.00 & 10.00 & 43.50 & 45.11
& 70.38 \\
\textbf{+DTR (Ours)}
& \best{42.98} & \best{50.78} & \best{82.35} & \best{56.48}
& 50.50 & \best{68.18} & \best{56.00} & 11.50 & \best{48.00} & \best{46.84}
& 70.55 \\
\bottomrule
\end{tabular}
\end{adjustbox}
\end{table*}

As shown earlier in the paper, DTR improves hallucination robustness across LLaVA- and Qwen-based backbones while preserving competitive video understanding performance. All results reported below use the same 8-frame sampling protocol as used earlier in the paper. For Qwen-VL models, we report VideoHallucer accuracy on hallucination-targeted questions, consistent with the evaluation protocol used earlier in the paper. Here we further report results on Qwen3-VL-4B~\cite{bai2025qwen3}, InternVL3.5-8B, and InternVL3.5-14B~\cite{wang2025internvl3} to further demonstrate the effectiveness of \method{} across additional backbones with different scales and architectural characteristics. Taken together, the results in Tables~\ref{tab:qwen3vl_4b_results}--\ref{tab:internvl35_14b_results} further support the generality of \method{} across model families and scales.

On Qwen3-VL-4B, \method{} improves EventHallusion overall from 66.01 to 71.15 and VideoHallucer from 79.68 to 82.36. These gains suggest that decoder-side temporal rebalancing remains effective even on a smaller backbone.

On InternVL3.5-8B, \method{} improves EventHallusion overall from 53.79 to 58.92 and VideoHallucer overall from 48.98 to 50.08. This pattern suggests that the benefit of \method{} is not limited to Qwen- or LLaVA-based models, but also extends to InternVL backbones with different architectural characteristics.

On InternVL3.5-14B, \method{} further improves EventHallusion overall and VideoHallucer overall. Compared with the 8B variant, this result suggests that the effectiveness of decoder-side temporal rebalancing generalizes across model scales.

\section{Overall Summary}

The results in this appendix further strengthen the main conclusion in three aspects. First, they show that the observed anchor-frame bias and the effect of \method{} remain stable under different frame-sampling densities, indicating that the phenomenon is not tied to a particular sampling setup. Second, they show that similar temporal selection bias also appears in model families beyond LLaVA, although its exact pattern varies across architectures, suggesting that the problem is more general. Third, they further demonstrate the effectiveness of DTR across additional backbones and model scales, supporting its robustness and generality. Overall, the appendix strengthens the view that mitigating hallucinations in Video-LLMs requires correcting temporal evidence imbalance on the decoder side, which is exactly what \method{} is designed to do.

\end{document}